# LLMs-Powered Accurate Extraction, Querying and Intelligent Management of Literature-derived 2D Materials Data

Lijun Shang[1,2], Yadong Yu[1,*], Wenqiang Kang[1,2], Jian Zhou[3], Dongyue Gao[2], Pan Xiang[1], Zhe Liu[1], Mengyan Dai[1,*], Zhonglu Guo[2], Zhimei Sun[3,*]

[1]Chemical Defense Institute, Academy of Military Sciences, Beijing 102205, China.

[2]Hebei Key Laboratory of Boron Nitride Micro and Nano Materials, School of Materials Science and Engineering, Hebei University of Technology, Tianjin 300130, China.

[3]School of Materials Science and Engineering, and Center for Integrated Computational Materials Science, International Research Institute for Multidisciplinary Science, Beihang University, Beijing 100191, P. R. China.

*Correspondence and requests for materials should be addressed to Y. D. Yu or M. Y. Dai or Z. M. Sun: ydyu16@buaa.edu.cn, daidecai0558@163.com, zmsun@buaa.edu.cn.

# Abstract

Two-dimensional (2D) materials have showed widespread applications in energy storage and conversion owning to their unique physicochemical, and electronic properties. Most of the valuable information for the materials, such as their properties and preparation methods, is included in the published research papers. However, due to the dispersion of synthetic sources and the inconsistency of performance reported in the literature, the data in the literature is difficult to be reused, thereby limiting the further research and progress of 2D materials. To this end, we proposed an end-to-end framework that couples large language models (LLMs) with domain informed context engineering and lightweight fine-tuning to extract structured knowledge at large-scale from the published papers, and then the knowledge is available externally through an agent-assisted data management system. As a result, the literature is collected via programmatic discovery and converted to standardized plain text, where the synthesis details, performance metrics, and article metadata are all included. We have applied the workflow to about 50,000 papers, curating 600,200 performance and 202,300 synthesis records. Compared with prompt-only extraction, context engineering has significantly improved the precision and recall, and fine-tuning can further increase the coverage of domain terminology. These data have been structured into a relational knowledge base with natural-language retrieval and guarded SQL execution. It achieves nearly perfect accuracy in simple and moderately complex queries, and can also achieve a relatively high accuracy even in complex multi-table queries. This work demonstrates how LLMs accelerate dataset construction, enable reproducible benchmarking, and support synthesis pathway design for 2D materials.

# 1. Introduction

In recent years, Two-dimensional (2D) materials have garnered significant interest due to their unique properties including physical, chemical, and electronic properties[1-5]. To date, a variety of 2D materials such as 2D transition metal carbides and nitrides (MXenes), 2D transition metal borides (MBenes), transition metal dichalcogenides (TMDs) and hexagonal boron nitride (hBN) have been discovered and successfully employed in various applications including energy storage and conversion and $CO_2$ reduction catalysis, flexible electronics and assorted photonic devices, and seawater desalination and water-purification membranes for environmental remediation[6-10]. However, in most cases, 2D materials data are limited in practical use due to the lack of complete synthetic provenance and high-quality characterization datasets[11, 12]. To further amplify 2D materials dataset. Campi et al. expanded the Materials Cloud 2D database under an extended high-throughput exfoliation-screening framework with DFT validation, obtaining 3,077 monolayers in total, including 2,004 that are easily exfoliable. Similarly, Momeni et al. mentioned that utilizing high-throughput screening and integrating experimental validation methods to expand the dataset for 2D materials[13, 14]. Extensive studies in this domain are mainly focused on employing high-throughput computational screening or experimental synthesis to augment datasets, but few reports have described expanding 2D materials datasets by mining data from published literature[15, 16], which possess a large amount of data. Compared with the time and resource intensive high-throughput computations or experimental syntheses, directly extracting validated results from previously published studies offers faster and more accurate 2D materials data. In the previous studies, Vaucher et al. built a transformer sequence-to-sequence model to translate prose experimental procedures into structured, automation-ready action sequences. Pretrained on rule-based NLP outputs and refined with manual annotations, it achieved 60.8 percent exact sequence matches on a challenging test set; at thresholds of at least 90 percent and at least 75 percent, the match rates were 71.3 percent and 82.4 percent, respectively, demonstrating strong extraction precision for organic synthesis[17]. Tshitoyan et al. trained two-hundred-dimensional Word2vec embeddings on approximately 3.3 million materials-science abstracts for unsupervised literature mining, capturing periodic trends and structure–property relations. Using cosine similarity between material vectors and application keywords (for example, "thermoelectric"), the model retrospectively surfaced candidates not previously linked to the application: the top 50 predictions were approximately eight times more likely to be studied within five years than randomly selected unstudied materials, and the rankings showed Spearman correlations of 0.59 with power factor and 0.52 with zT[18]. Jang et al. proposed a semi-supervised text mining method to extract synthesis and processing data, focusing on action sequences and parameters and using a small labeled corpus combined with unlabeled text to build a robust superalloy dataset[19]. Therefore, combining literature-extracted synthesis parameters with performance data for 2D materials into a dataset can improve data quality while expanding the database. However, most reported text-mining methods are restricted to domain-specific corpora. Additionally, the canonicalization pipelines required to standardize the extracted data are markedly time-consuming and inefficient[20, 21]. Even in the simplest case, all extraction parameters must be optimized simultaneously. The search-and-optimization cycle must then be restarted from scratch for each target attribute. Meanwhile, extraction precision depends on researchers fully understanding the data, which greatly slows the process. Therefore, building an on-demand method that pulls information from all materials-science papers remains a major challenge.

Nowadays, large language models (LLMs), propelled by rapid advances in artificial intelligence, are gaining significant attention as a powerful and flexible means of tackling the above issues[22]. In earlier studies, Zheng et al. developed a workflow utilizing ChatGPT as a collaborator for human chemists, extracting 26,257 distinct synthesis parameters of approximately 800 metal-organic frameworks from 228 articles[23]. Polak and Morgan proposed a similar workflow for metallic glasses and high entropy alloys, employing follow-up questions to GPT-4 to ensure correctness and address the issues of hallucinations with LLMs[24]. Yang et al. used a repeated questioning strategy with GPT-4 for bandgap values, demonstrating reduced error rates and a more extensive dataset than human-curated databases[25]. Similarly, Schmidt et al. systematically assessed LLMs for data extraction and confirmed their feasibility. However, LLMs achieve only about 70% precision on structured data extraction tasks within specific domains; on open-source small models, this figure is often even lower[26]. To confront this limitation, task-specific fine-tuning has come to the fore[27]. Wang et al. fine-tuned LLMs, enhancing their performance in chemical text mining tasks like compound recognition and reaction role labeling, with minimal data and high efficiency[28]. By applying fine-tuning techniques to LLMs and leveraging the rich literature on 2D materials, LLMs can rapidly extract and organize synthesis and performance data. In addition, the code generation ability of LLMs makes them promising methods for database management. Therefore, it is worth noting that combining LLMs with domain-specific adaptation can markedly enhance the speed and precision of 2D materials database construction and upkeep. Yet a full end-to-end workflow—from information extraction to database management—powered by LLMs has not yet been described to date.

In this study, we have investigated and integrated the LLM-powered 2D material information extraction technology to our proprietary data intelligent management platform. It enabled the construction of an application framework centered around an LLM agent, thereby achieving rapid retrieval, convenient access, and systematic analysis of 2D material information (as shown in Fig. 1). To begin with, literature is collected via OpenAlex using 2D material keywords or curated DOI lists to obtain download links (Fig. 1a). The collected corpus is compiled into a 2D materials text database and segmented by the Segment Any Text (SaT) model, producing standardized input files[29]. Notably, 3 000 manually annotated samples are used to fine-tune a model dedicated to extracting 2D materials information. In addition, in view of prompt sensitivity, context engineering has been introduced to allowing the model to utilize structured cues to precisely retrieve synthesis and property data, thereby improving extraction precision[30]. Finally, step-by-step extraction of article titles, Digital Object Identifiers (DOIs), synthesis details, and performance data is achieved by pairing the fine-tuned model with tailored prompts and structured input files[31] (Fig. 1b). By combining LLM fine-tuning on limited labeled data with prompt engineering, we have successfully resolved the performance degradation encountered when deploying smaller models for information extraction in the 2D materials domain. The performance of the extracted results in extraction precision has approached top-tier open-source or closed-source models such as DeepSeek V3, Qwen3-235B-A22B, and Gemini 2.5 Flash[32, 33]. For effective data management, search, and analysis, the LLM-based data intelligent management system has integrated a set of intelligent agents that let users retrieve information from the database using natural language. The system also learns over time: repeated queries deepen its understanding of the database structure, steadily improving the precision and sophistication of data extraction (Fig. 1c). This integrated framework provides a foundation for accelerating materials discovery, with potential extensions to other emerging material systems and integration with automated experimental workflows for closed-loop materials development.

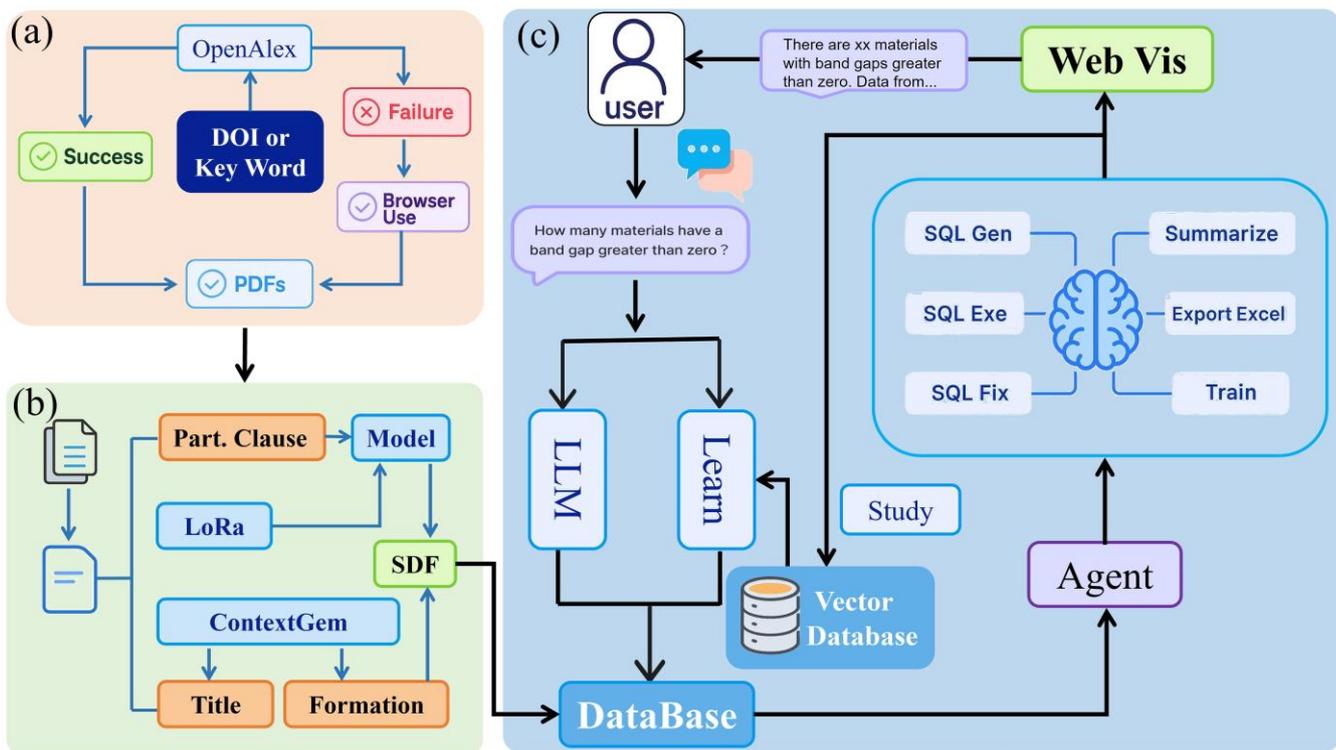

Fig. 1. Overall workflow of the literature collection and structured database construction . The pipeline comprises three stages: (a) literature retrieval and download; (b) literature processing and information extraction; and (c) the basic operating workflow of the data intelligent management system.

# 2. Results and discussion

## 2.1 Literature download and text structuring

In the literature download preparation stage, it is crucial to judge whether the literature's topic focuses on 2D materials. To ensure the accurate downloading of relevant literature, we first collected over 134,000 literature entries related to 2D materials, covering essential details such as authors, publication sources, publication years, and DOIs. As shown in Fig. 2a, first, OpenAlex was utilized to retrieve open-access paper download links based on the DOIs obtained above. This was supplemented by keyword-based searches to acquire additional relevant literature. Using a combination of link access and manual downloading, over 50,000 initially screened PDF documents related to 2D materials were obtained. To ascertain the validity of the corpus, we randomly sampled 1,000 documents and evaluated their relevance with three LLMs (GPT-4.1-mini, Gemini 2.5 Flash, and DeepSeek-R1). As shown in Fig. 2b, over 95% of the sampled documents were judged to be related to 2D materials. Further analysis (Fig. 2c) showed that graphene constituted the largest share (37.5%), followed by TMDs (23.8%) and MXenes (20%), whereas h-BN accounted for only 0.84%. Given possible errors in model classification, we kept all 50,000 documents to reduce false negatives for extraction and analysis. Subsequent, all collected PDFs were converted to text with MinerU, a document parsing tool designed for batch automated content extraction[34, 35]. The extracted texts were then standardized using the SaT model, which employs segmentation algorithms combined with semantic understanding to identify and normalize paragraph and sentence boundaries based on contextual meaning. Finally, this structured text-processing approach not only enables more accurate identification of target sentences or paragraphs during subsequent information extraction but also precisely indicates the exact positions of the extracted information within the corresponding sentences when results are returned. Through structured processing of the text, preparations are made for subsequent 2D material information extraction and the construction of a data intelligent management system.

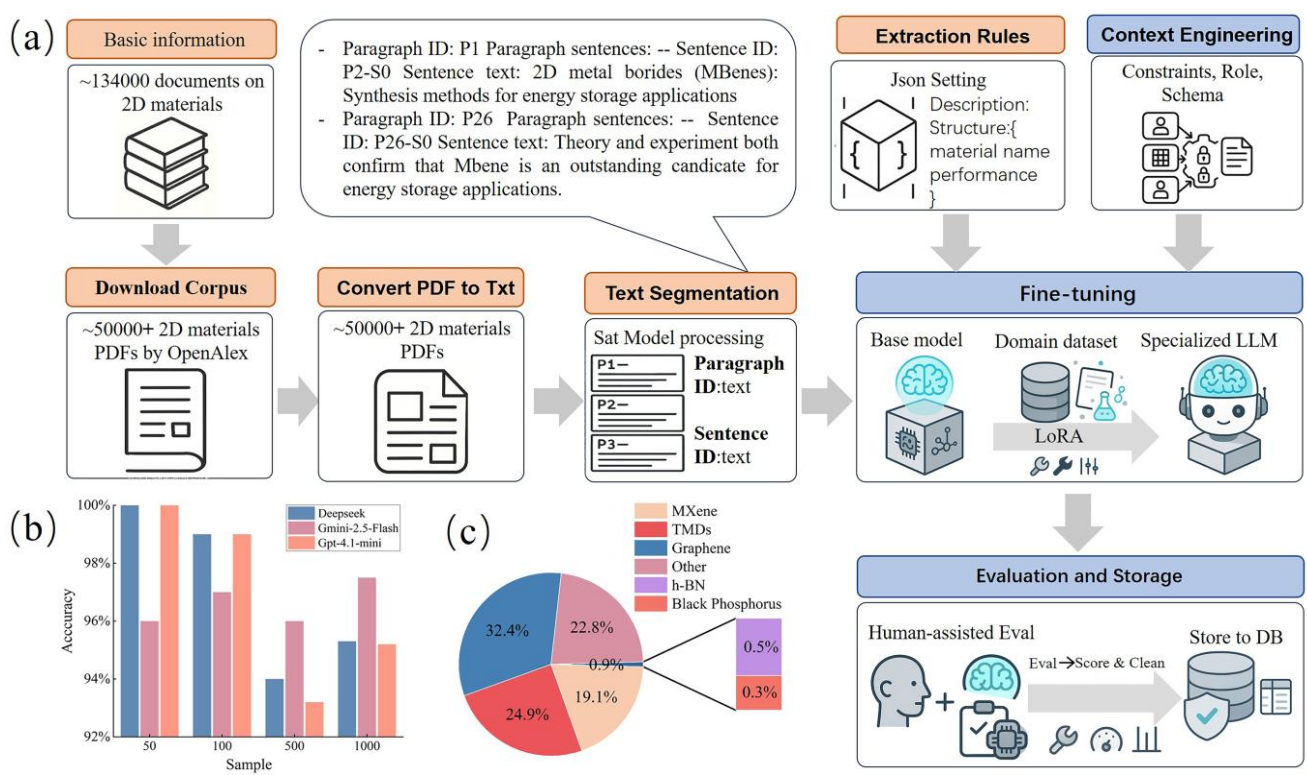

Fig. 2. Pipeline and results for processing 2D materials literature.(a) End-to-end workflow: download the corpus from OpenAlex; convert PDFs to text and perform sentence segmentation; construct a structured schema by combining extraction rules with context engineering; fine-tune the base model via LoRA; conduct human evaluation and then integrate into the database; (b) Bar chart showing the proportion of papers classified as 2D materials-related by three models (DeepSeek, Gemini-2.5-Flash, GPT-4.1-mini) across different sample sizes (50, 100, 500, 1,000); (c) Pie chart illustrating the distribution of material types identified in the literature corpus.

## 2.2 Data extraction using large language models

Information data on 2D materials from a journal article can be extracted by leveraging the ability of LLMs to understand the specialized semantics of materials entities discussed in the text. However, obtaining the desired output from an LLM poses a challenging task and is presently a subject of active research. Even with the same prompt or instruction and text generation parameters, the LLMs can produce varying responses. This variation is even greater across different model, making response quality hard to predict. Simple natural language prompts do not always produce the expected results from LLMs, because these models interpret language in non-intuitive ways. Hence, tightening contextual constraints and adopting techniques and parameter settings that enforce a consistent response format to reduce stochastic variability are crucial. Equally crucial is targeted fine-tuning of smaller models, so that in the specific 2D materials domain they can approach the response quality of stronger models.

To address these issues, we no longer relied only on prompt engineering. Prompt-only approaches are sensitive to wording and order, leading to inconsistent results across different documents and runs. They also fail to ensure that the output matches the required JSON schema. Domain-specific synonyms and rare cases are poorly handled, which increases errors and reduces information coverage. To improve stability and consistency, we used context engineering to assign the model a fixed role as an information extractor which was responsible for extracting synthesis details, article metadata such as title and DOI, and materials performance, and by adding a simple, uniform JSON schema and rules on bottom of the information extractor, we could efficiently and accurately extract specific information. This approach provides stricter control over the model's behavior while maintaining extraction flexibility, producing consistent, structured results that are easy to verify. To assess gains in structural completeness and extraction precision from context engineering, we randomly sampled 1,000 documents and used three models, DeepSeek-V3, Qwen3-235B-A22B and Gemini-2.5-Flash. Each model ran under two settings, prompt engineering and context engineering. For both settings, we used a bipartite matching method with a similarity threshold. The gold set is the human-annotated reference. TP is a predicted item that matches a gold item. FP is a predicted item that matches none. FN is a gold item that is not matched by any prediction. As shown in Fig. 3a, for text pairs, we used SequenceMatcher along with bigram-based and trigram-based Jaccard similarity to score each text block (see Supplementary Eq. (1) to (3))[36]. These scores were then combined through a weighted aggregation to produce the overall similarity for the pair. For numeric pairs, we computed similarity scores by comparing values directly, with the final score as a weighted sum of numeric components. When similarity exceeded a threshold (e.g., 0.65), we created an edge between gold and predicted items and applied one-to-one maximum matching to determine TP, FP, and FN, then calculated strict precision, recall, and F1. The one-to-one constraint penalizes outputs that merge several facts into one or segment too coarsely. In such cases a single prediction cannot cover multiple gold items, so recall and F1 may be underestimated. In Fig. 2b, context engineering markedly improves extraction success for both open-source models such as DeepSeek-V3 and closed-source models such as Gemini 2.5 Flash, reducing failure rates from about 30% to below 5% (as show in Supplementary Fig. S1). In addition to reliability gains, precision and recall also increase. For DeepSeek-V3, precision rises from 64% to 91% (up 27 percentage points, a 42% relative increase), and recall rises from 57% to 67% (up 10 percentage points, a 17% relative increase). These results indicate that context engineering is effective for materials information extraction.

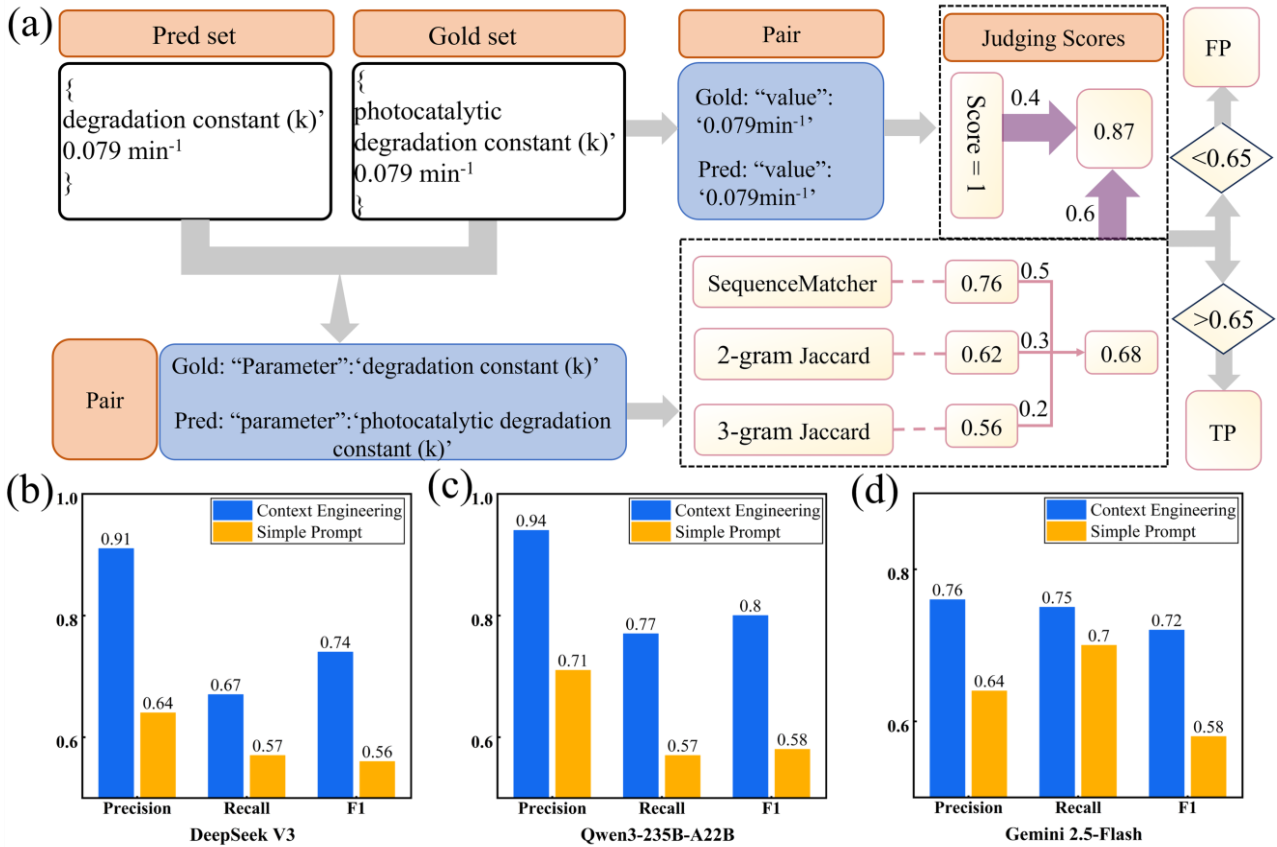

Fig. 3. Evaluation workflow and comparative results. (a) Schematic of the evaluation workflow for model-based information extraction;(b)–(d) Bar charts of the evaluation results (Precision, Recall, F1) for DeepSeek V3, Qwen3-235B-A22B, and Gemini-2.5-Flash under two settings: simple prompts and context engineering.

To push a smaller model toward or beyond higher-tier performance, we applied targeted adjustments. Foundation models freeze parameters after pretraining and cover limited domain knowledge, for example confusing MXenes or TMDs with specific materials, which compromises extraction quality. We curated a training dataset of 3,000 manually annotated examples and expanded it to 10,000 samples through paraphrase augmentation to enhance data diversity while maintaining annotation quality. For 2D-materials extraction task we chose parameter-efficient LoRA rather than full fine-tuning because our domain corpus is much smaller than the pretraining data, and full updates risk degrading existing abilities and raising memorization variants during fine-tuning, which reduces terminology ambiguity[37]. As shown in Fig. 4b, we monitored both training and validation losses throughout the fine-tuning process. The validation loss exhibited a minimum at 1,500 training steps, after which it demonstrated an upward trend, signaling the onset of overfitting. Consequently, we selected the checkpoint at step 1,500 as our final extraction model to ensure optimal generalization performance. The fine-tuned model, combined with structural constraints from prompt engineering, improves field detection and hierarchical structures, achieving schema accuracy and extraction performance close to leading models. As illustrated in Fig. 4c, this approach achieved a precision of 0.83 and a recall of 0.66, demonstrating substantial improvements over the baseline. Subsequently, we then used the fine-tuned model to process over fifty thousand papers in batch. The results kept high output quality while reducing cost versus commercial models. Fig. 4d shows that extracting from two thousand papers costs \$26.78 with DeepSeek V3, \$23.79 with Qwen3-235B-A22B, and \$87.68 with Gemini-2.5-Flash. By contrast, our smaller model runs locally on a single RTX 4090, reducing total cost by about 90% while avoiding API-related data exposure. Finally, we extracted data from 50,000 papers and, prior to storage, ensured precision, stability, and utility through manual review and the removal of low-scoring or anomalous outputs, yielding a curated set of 600,200 performance records and 202,300 synthesis records, for which Supplementary Tables 1 and 2 present a subset. Based on this, we constructed a computable knowledge base for 2D materials. In an era where data value is paramount, this knowledge base can be used to fine-tune large models in the materials field or to build dedicated data foundations through Retrieval Augmented Generation (RAG)[38]. This approach enables rational design of synthesis pathways, parameter optimization, and accelerated discovery of new materials. Supported by high-quality, curated data, machine-learning models achieve improved generalization and transferability, while standardized experimental protocols exhibit greater comparability and reproducibility. Together, data and models shift materials research from experience-driven practice to a data-guided workflow, transforming fragmented literature into structured, computable, and actionable data assets.

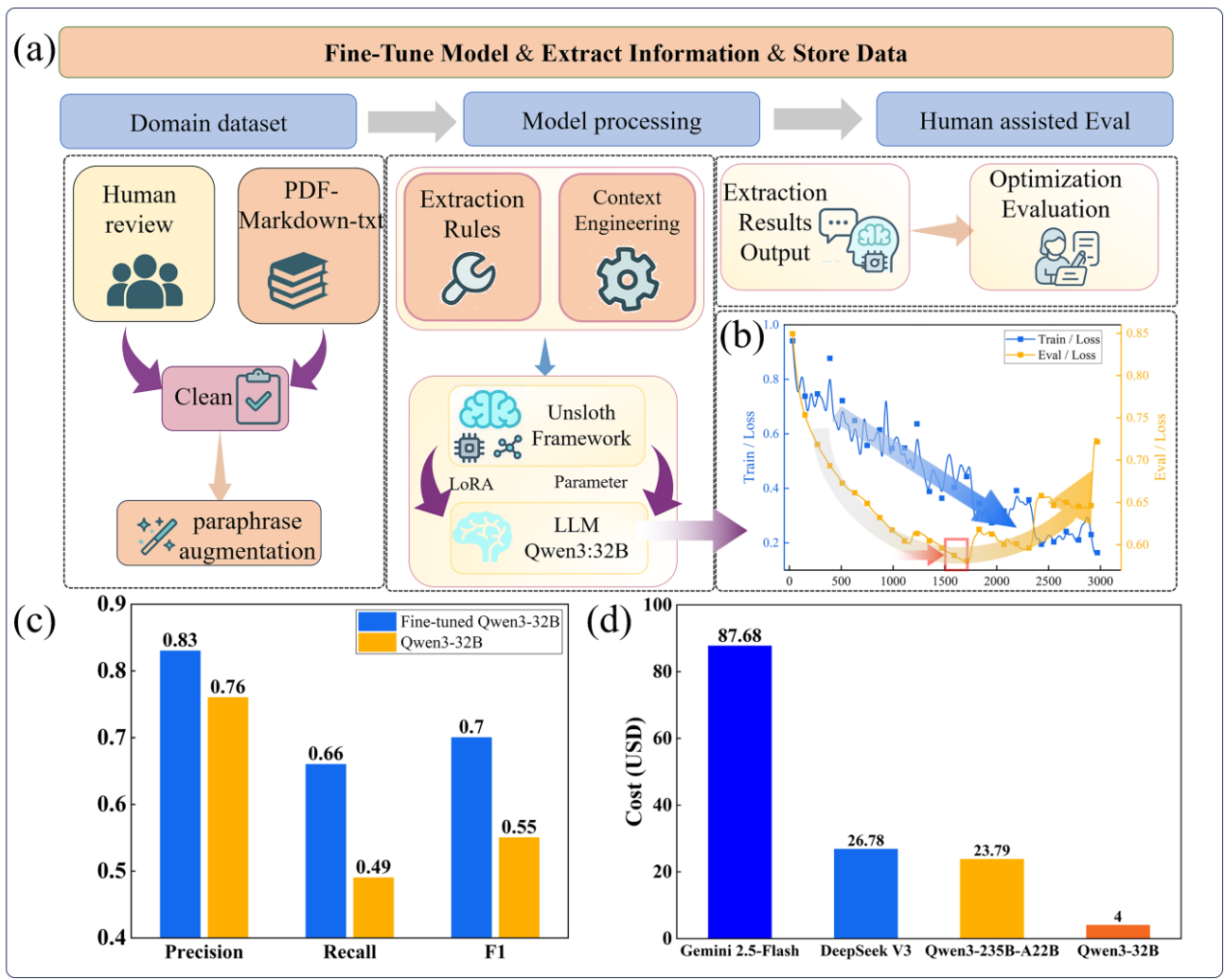

Fig. 4. (a) Schematic of the end-to-end pipeline for model fine-tuning; (b) Training and evaluation curves during fine-tuning; (c) Comparative evaluation of the fine-tuned model versus the base model (Precision, Recall, F1); (d) Cost of extracting information from 2,000 papers across different models.

## 2.3 Data Intelligent Management

Data management and use are as important as data extraction. To ensure long-term maintainability and reproducibility of performance and synthesis information on 2D materials, we store the extracted results in a relational database[39]. The backend uses MySQL and provides a stable, general foundation for downstream search and analysis[40]. For natural-language queries, we have built a multi-agent data intelligence system as shown in Fig. 5. When a user submits a query, the routing agent analyzes the intent and forwards the request to a SQL-generation agent to construct the statement. A safety agent reviews the SQL and blocks delete, truncate, and overwrite operations to protect the database boundary. Approved statements are sent to an execution agent to access the database. The execution process includes retry logic and error handling. If execution fails due to syntax errors or constraint violations, the error message is fed back to the generation agent, which regenerates and validates the SQL[41]. If the failure persists, a repair agent refines the SQL statement using contextual information and error diagnostics until successful execution is achieved. A validation agent then verifies that the query results align with the original user intent. Finally, a summarization agent records the query and result for audit and decides whether to export to Excel by information granularity and use. Aggregate requests return a number, for example how many materials have a band gap greater than zero. Detailed requests provide a downloadable file, for example the records of materials with a band gap greater than zero. The system stays flexible in model choice. Each agent selects a suitable model by task difficulty to balance quality and cost. Higher-capability models are prioritized for SQL generation and repair tasks, while lightweight models handle parsing and summarization. Models can be dynamically swapped to optimize for workload and cost constraints. All agent interfaces maintain compatibility with OpenAI API conventions. For data and privacy protection, we recommend local deployment via LM Studio, Ollama, or vLLM. This supports a closed loop from question to database query, and from result checking to output packaging, without sending data outside.

## 2.4 Intelligent Learning for Databases

Database query quality depends on model capability. However, purely model-based methods face challenges with diverse phrasings, unclear field meanings, and complex logic—resulting in inconsistent performance and query errors. We propose a query-focused learning framework with active and passive modes. In active mode, the language model evaluates the query's importance and learning value, then automatically incorporates mappings between the query text, key fields, and validated results into the example repository. In passive mode, users retain discretion over whether to add the query mappings after execution. As examples accumulate, the system progressively deepens its understanding of tables and fields. Consequently, intent recognition accelerates, ambiguity diminishes, and both accuracy and consistency improve, while the need for repeated parsing and manual corrections decreases. This feature preserves the existing workflow while capturing reusable knowledge from each query execution, thereby reducing token consumption and enhancing both speed and accuracy for subsequent similar queries (as illustrated in Supplementary Fig. S2). To evaluate the accuracy of materials data queries, we categorized questions into three levels of complexity: simple, medium, and complex, with 100 questions per category. Simple queries involved straightforward counting tasks, such as determining the number of instances of a specific material or the total database size. Medium-complexity queries extended these basic searches by requiring data export functionality, necessitating more sophisticated SQL workflows. Complex queries spanned multiple tables and demanded comprehensive schema knowledge, incorporating aggregation operations, fundamental calculations, and data sorting. The system achieved near-perfect accuracy (approaching 100%) on simple and medium-complexity tasks, and approximately 90% accuracy on complex tasks. This materials-focused system maintains high accuracy in data querying and extraction while significantly reducing access barriers. It delivers analysis-ready, high-quality data for synthesis pathways design, performance evaluation, and mechanistic studies, while enabling materials-centric intelligent information retrieval through intuitive human-computer interaction.

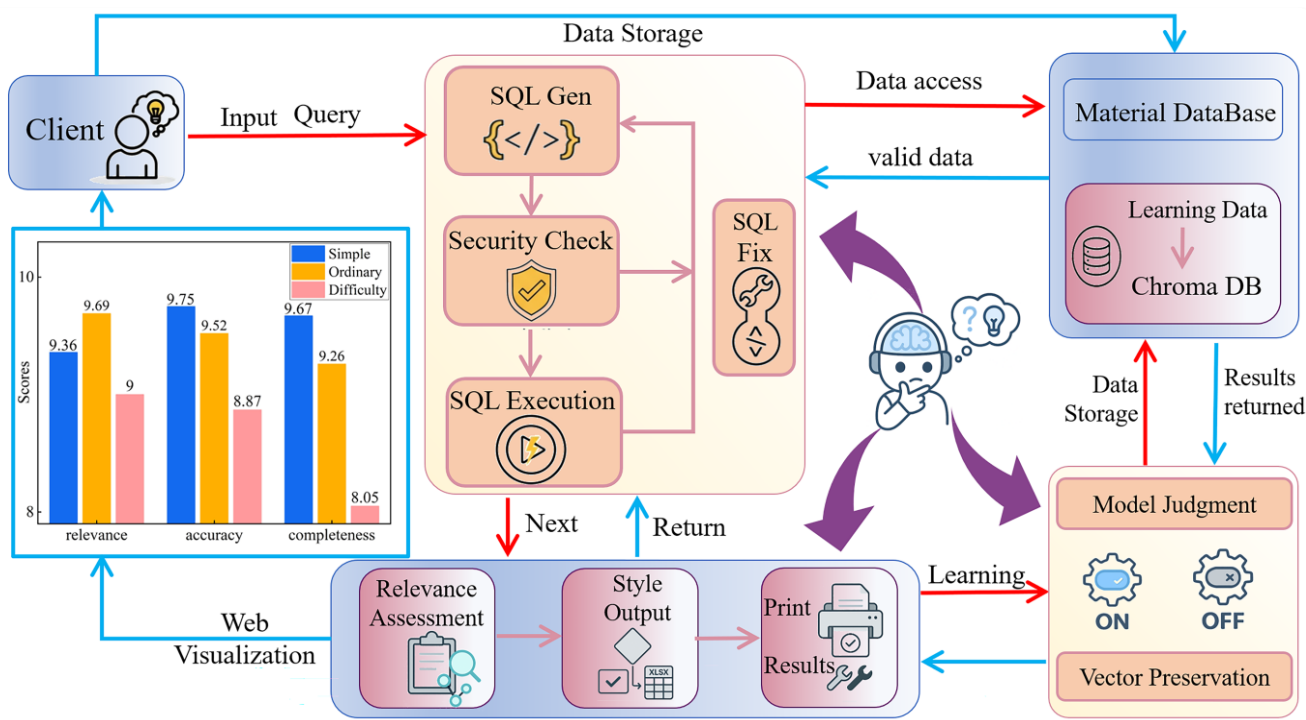

**Fig. 5.** Workflow for constructing the agent in the data intelligent management system, and extraction accuracy across 100 tasks at three difficulty levels—simple, ordinary, and difficult.

## 3. Conclusion

This work has built a structured database of 2D materials and an intelligent search and analysis platform on materials data, achieving unified standardization and traceable management of three categories of information: "structure–performance–synthesis." The results show that by combining small-sample fine-tuning for a specific domain with context engineering can significantly improve the stability and reusability of extracting 2D-materials information from the literature. Based on more than 50,000 papers, we have collected about 600,000 performance records and 202,000 synthesis records to construct a relational knowledge base, and have implemented an auditable, traceable, and secure retrieval and analysis capability via a multi-agent natural-language-to-SQL layer. Overall, this work provides a solid data foundation for optimizing synthesis pathways, conducting cross-materials-system benchmarking, and generating new hypotheses.

To further improve performance, two key aspects will be addressed in the next phase of work. First, the matching strategies in the evaluation framework will be refined to address the low recall rates caused by matching limitations, thereby more accurately reflecting the actual system performance. Second, the scope of information extraction will be expanded to encompass not only textual content but also tables and figures from the literature through systematic extraction and structured processing, enabling more comprehensive knowledge acquisition.

## Declaration of competing interest

The authors declare that they have no known competing financial interests or personal relationships that could have appeared to influence the work reported in this paper.

## CRediT authorship contribution statement

**Lijun Shang**: conducted the main research work, implemented the code, and wrote the manuscript; **Yadong Yu**: proposed the core research idea and provided technical guidance; **Wenqiang Kang**: and **Dongyue Gao**: assisted with manuscript writing and participated in the design of figures; **Mengyan Dai**: Writing –review & editing, Validation, Resources, Investigation; **Pan Xiang** and **Zhe Liu**: provided support for the code testing environment; **Zhonglu Guo**: conducted system testing and provided technical guidance; **Jian Zhou** and **Zhimei Sun**: provided funding support and steer the primary research directions.

## Acknowledgements

This work was supported by the National Natural Science Foundation of China (No. 52403306) and the Hebei Natural Science Foundation (No. B2024202047).

## Ethical approval and consent to participate

Not applicable.

## Code availability

All data and code of this study are publicly available at the GitHub repository https://github.com/Moneshanghai/database The repository includes end to end code for extracting corpora from two dimensional materials papers a data intelligence system for organization retrieval and visualization and the extracted dataset approximately six hundred thousand performance records and more than two hundred thousand synthesis records. Further inquiries can be directed to the corresponding author(s).

# References

[1]    S. Hong, F. Al Marzooqi, J.K. El-Demellawi, N. Al Marzooqi, H.A. Arafat, H.N. Alshareef, ACS Materials Letters, 5 (2023) 341-356. DOI: 10.1021/acsmaterialslett.2c00914

[2]    D. Jiang, Z. Liu, Z. Xiao, Y. Qian, Y. Sun, Z. Zeng, R. Wang, J. Mater. Chem. A, 10 (2022) 89-121. DOI: 10.1039/D1TA06741A

[3]    A.J. Khan, S.S. Shah, S. Khan, A. Mateen, B. Iqbal, M. Naseem, L. He, Y. Zhang, Y. Che, Y. Tang, M. Xu, L. Gao, G. Zhao, Chem. Eng. J., 497 (2024) 154429. DOI: 10.1016/j.cej.2024.154429

[4]    P. Aghamohammadi, B. Hüner, O.C. Altıncı, E.T. Akgul, B. Teymur, U.B. Simsek, M. Demir, Int. J. Hydrogen Energy, 87 (2024) 179-198. DOI: 10.1016/j.ijhydene.2024.08.412

[5]    X. Wang, N.-D. Tran, S. Zeng, C. Hou, Y. Chen, J. Ni, npj Comput. Mater., 8 (2022) 253. DOI: 10.1038/s41524-022-00945-x

[6]    R. Ali, M. Islam, M. Shafi, S. Ali, H.-E. Wang, Coord. Chem. Rev., 540 (2025) 216797. DOI: 10.1016/j.ccr.2025.216797

[7]    S. Ogawa, S. Fukushima, M. Shimatani,  Materials2023.

[8]    Q.H. Wang, K. Kalantar-Zadeh, A. Kis, J.N. Coleman, M.S. Strano, Nat. Nanotechnol., 7 (2012) 699-712. DOI: 10.1038/nnano.2012.193

[9]    T. Xu, Y. Wang, Z. Xiong, Y. Wang, Y. Zhou, X. Li, Nano-Micro Letters, 15 (2022) 6. DOI: 10.1007/s40820-022-00976-5

[10]   H. Meskher, A.K. Thakur, S.K. Hazra, M.S. Ahamed, A.M. Saleque, Q.F. Alsalhy, M.W. Shahzad, M.N.A.S. Ivan, S. Saha, I. Lynch, Environmental Science: Nano, 12 (2025) 1012-1036. DOI: 10.1039/D4EN00427B

[11]   M.J. Statt, B.A. Rohr, D. Guevarra, S.K. Suram, T.E. Morrell, J.M. Gregoire, Scientific Data, 10 (2023) 184. DOI: 10.1038/s41597-023-02107-0

[12]   U. Celano, D. Schmidt, C. Beitia, G. Orji, A.V. Davydov, Y. Obeng, Nanoscale Advances, 6 (2024) 2260-2269. DOI: 10.1039/d3na01148h

[13]   D. Campi, N. Mounet, M. Gibertini, G. Pizzi, N. Marzari, ACS Nano, 17 (2023) 11268-11278. DOI: 10.1021/acsnano.2c11510

[14]   K. Momeni, Y. Ji, Y. Wang, S. Paul, S. Neshani, D.E. Yilmaz, Y.K. Shin, D. Zhang, J.-W. Jiang, H.S. Park, S. Sinnott, A. van Duin, V. Crespi, L.-Q. Chen, npj Comput. Mater., 6 (2020) 22. DOI: 10.1038/s41524-020-0280-2

[15]   R. Zhang, J. Zhang, Q. Chen, B. Wang, Y. Liu, Q. Qian, D. Pan, J. Xia, Y. Wang, Y. Han, Computational Materials Science, 230 (2023) 112441. DOI: 10.1016/j.commatsci.2023.112441

[16]   M. Yao, J. Ji, X. Li, Z. Zhu, J.-Y. Ge, D.J. Singh, J. Xi, J. Yang, W. Zhang, Science China Materials, 66 (2023) 2768-2776. DOI: 10.1007/s40843-022-2401-3

[17]   V. Tshitoyan, J. Dagdelen, L. Weston, A. Dunn, Z. Rong, O. Kononova, K.A. Persson, G. Ceder, A. Jain, Nature, 571 (2019) 95-98. DOI: 10.1038/s41586-019-1335-8

[18]   A.C. Vaucher, F. Zipoli, J. Geluykens, V.H. Nair, P. Schwaller, T. Laino, Nat. Commun., 11 (2020) 3601. DOI: 10.1038/s41467-020-17266-6

[19]   W. Wang, X. Jiang, S. Tian, P. Liu, T. Lookman, Y. Su, J. Xie, npj Comput. Mater., 9 (2023) 183. DOI: 10.1038/s41524-023-01138-w

[20]    O. Kononova, H. Huo, T. He, Z. Rong, T. Botari, W. Sun, V. Tshitoyan, G. Ceder, Scientific Data, 6 (2019) 203. DOI: 10.1038/s41597-019-0224-1

[21]    T. Gupta, M. Zaki, N.M.A. Krishnan, Mausam, npj Comput. Mater., 8 (2022) 102. DOI: 10.1038/s41524-022-00784-w

[22]    S. Lappin, Journal of Logic, Language and Information, 33 (2024) 9-20. DOI: 10.1007/s10849-023-09409-x

[23]    Z. Zheng, O. Zhang, C. Borgs, J.T. Chayes, O.M. Yaghi, J. Am. Chem. Soc., 145 (2023) 18048-18062. DOI: 10.1021/jacs.3c05819

[24]    M.P. Polak, D. Morgan, Nat. Commun., 15 (2024) 1569. DOI: 10.1038/s41467-024-45914-8

[25]    S. Yang, S. Li, S. Venugopalan, V. Tshitoyan, M. Aykol, A. Merchant, E. Cubuk, G. Cheon, arXiv (2023). DOI: 10.48550/arXiv.2311.13778

[26]    L. Schmidt, K. Hair, S. Graziosi, F. Campbell, C. Kapp, A. Khanteymoori, D. Craig, M. Engelbert, J. Thomas, arXiv, (2024). DOI: 10.48550/arXiv.2405.14445

[27]    T. Brown, B. Mann, N. Ryder, M. Subbiah, J.D. Kaplan, P. Dhariwal, A. Neelakantan, P. Shyam, G. Sastry, A. Askell, Adv. Neural Inf. Process. Syst., 33 (2020) 1877-1901. DOI: 10.48550/arXiv.2005.14165

[28]    W. Zhang, Q. Wang, X. Kong, J. Xiong, S. Ni, D. Cao, B. Niu, M. Chen, Y. Li, R. Zhang, Chemical science, 15 (2024) 10600-10611. DOI: 10.1039/d4sc00924j

[29]    M. Frohmann, I. Sterner, I. Vulić, B. Minixhofer, M. Schedl, arXiv, (2024). DOI: 10.48550/arXiv.2406.16678

[30]    L. Mei, J. Yao, Y. Ge, Y. Wang, B. Bi, Y. Cai, J. Liu, M. Li, Z.-Z. Li, D. Zhang, arXiv (2025). DOI: 10.48550/arXiv.2507.13334

[31]    J. Dagdelen, A. Dunn, S. Lee, N. Walker, A.S. Rosen, G. Ceder, K.A. Persson, A. Jain, Nat. Commun., 15 (2024) 1418. DOI: 10.1038/s41467-024-45563-x

[32]    J. Bai, S. Bai, Y. Chu, Z. Cui, K. Dang, X. Deng, Y. Fan, W. Ge, Y. Han, F. Huang, arXiv, (2023). DOI: 10.48550/arXiv.2309.16609

[33]    A. Liu, B. Feng, B. Xue, B. Wang, B. Wu, C. Lu, C. Zhao, C. Deng, C. Zhang, C. Ruan, arXiv (2024). DOI: 10.48550/arXiv.2412.19437

[34]    C. He, W. Li, Z. Jin, C. Xu, B. Wang, D. Lin, arXiv (2024). DOI: 10.48550/arXiv.2407.13773

[35]    J. Niu, Z. Liu, Z. Gu, B. Wang, L. Ouyang, Z. Zhao, T. Chu, T. He, F. Wu, Q. Zhang, arXiv (2025). DOI: 10.48550/arXiv.2509.22186

[36]    S. Garg, M. Vankadari, M. Milford,   Conference on Robot Learning, PMLR2022, pp. 429-443.

[37]    E.J. Hu, Y. Shen, P. Wallis, Z. Allen-Zhu, Y. Li, S. Wang, L. Wang, W.J.I. Chen, arXiv, 1 (2022) 3. DOI: 10.48550/arXiv.2106.09685

[38]    Z. Guo, L. Xia, Y. Yu, T. Ao, C. Huang, arXiv (2024). DOI: 10.48550/arXiv.2410.05779

[39]    M.D. Wilkinson, M. Dumontier, I.J. Aalbersberg, G. Appleton, M. Axton, A. Baak, N. Blomberg, J.-W. Boiten, L.B. da Silva Santos, P.E. Bourne, Scientific data, 3 (2016) 1-9. DOI: 10.1038/sdata.2016.18

[40]    J. Zhou, L. Shen, M.D. Costa, K.A. Persson, S.P. Ong, P. Huck, Y. Lu, X. Ma, Y. Chen, H. Tang, Scientific data, 6 (2019) 86. DOI: 10.1038/s41597-019-0097-3

[41]    A. Askari, C. Poelitz, X. Tang, arXiv, 39 (2025) 23433-23441. DOI: 10.48550/arXiv.2406.12692

# Supplementary Information for: LLMs-Powered Accurate Extraction, Querying and Intelligent Management of Literature-derived 2D Materials Data

Lijun Shang[1,2], Yadong Yu[1,*], Wenqiang Kang[1,2], Jian Zhou[3], Dongyue Gao[2], Pan Xiang[1], Zhe Liu[1], Mengyan Dai[1,*], Zhonglu Guo[2], Zhimei Sun[3,*]

[1]Chemical Defense Institute, Academy of Military Sciences, Beijing 102205, China.

[2]Hebei Key Laboratory of Boron Nitride Micro and Nano Materials, School of Materials Science and Engineering, Hebei University of Technology, Tianjin 300130, China.

[3]School of Materials Science and Engineering, and Center for Integrated Computational Materials Science, International Research Institute for Multidisciplinary Science, Beihang University, Beijing 100191, P. R. China.

*Correspondence and requests for materials should be addressed to Y. D. Yu or M. Y. Dai or Z. M. Sun: ydyu16@buaa.edu.cn, daidecai0558@163.com, zmsun@buaa.edu.cn.

## Part A — Equations

$$\mathrm{ratio}(s,t) = \frac{2M}{|s| + |t|} \tag{1}$$

Here $s, t$ are the sequences (typically strings), $|s|, |t|$ are their lengths, and $M$ is the sum of the lengths of all longest, non-overlapping matching blocks between $s$ and $t$. The similarity lies in $[0,1]$ and is symmetric. We use this definition to quantify block-based sequence similarity.

$$J_2(s,t) = \frac{|G_2(s) \cap G_2(t)|}{|G_2(s) \cup G_2(t)|}, \quad G_2(s) = \{s[i]s[i+1] \mid i = 1, \dots, |s|-1\} \tag{2}$$

A string is decomposed into the set $G_2(\cdot)$ of adjacent two-character substrings, and the similarity is defined by the intersection-over-union; $|\cdot|$ denotes set cardinality and $\cap, \cup$ are set intersection and union.

$$J_3(s,t) = \frac{|G_3(s) \cap G_3(t)|}{|G_3(s) \cup G_3(t)|}, \quad G_3(s) = \{s[i]s[i+1]s[i+2] \mid i = 1, \dots, |s|-2\} \tag{3}$$

Analogously, $G_3(\cdot)$ is the set of adjacent three-character substrings; this measure emphasizes longer contiguous matches.

## Part B — Figures

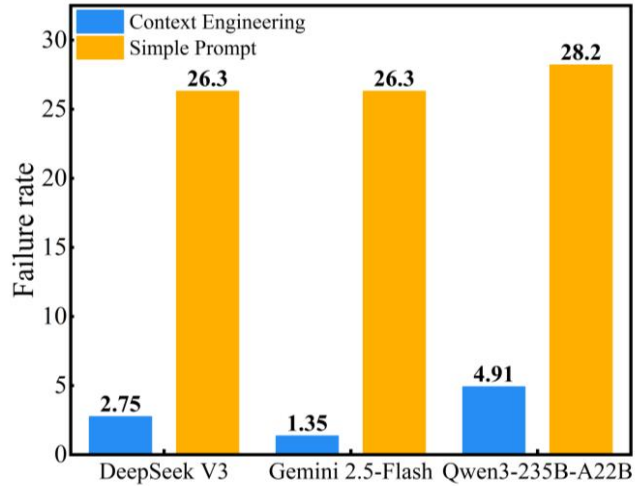

**Figure S1.** Failure rates for DeepSeek V3, Gemini 2.5-Flash, and Qwen3-235B-A22B under two prompting strategies. Context Engineering (blue) consistently reduces failures compared with Simple Prompt (orange) across all models.

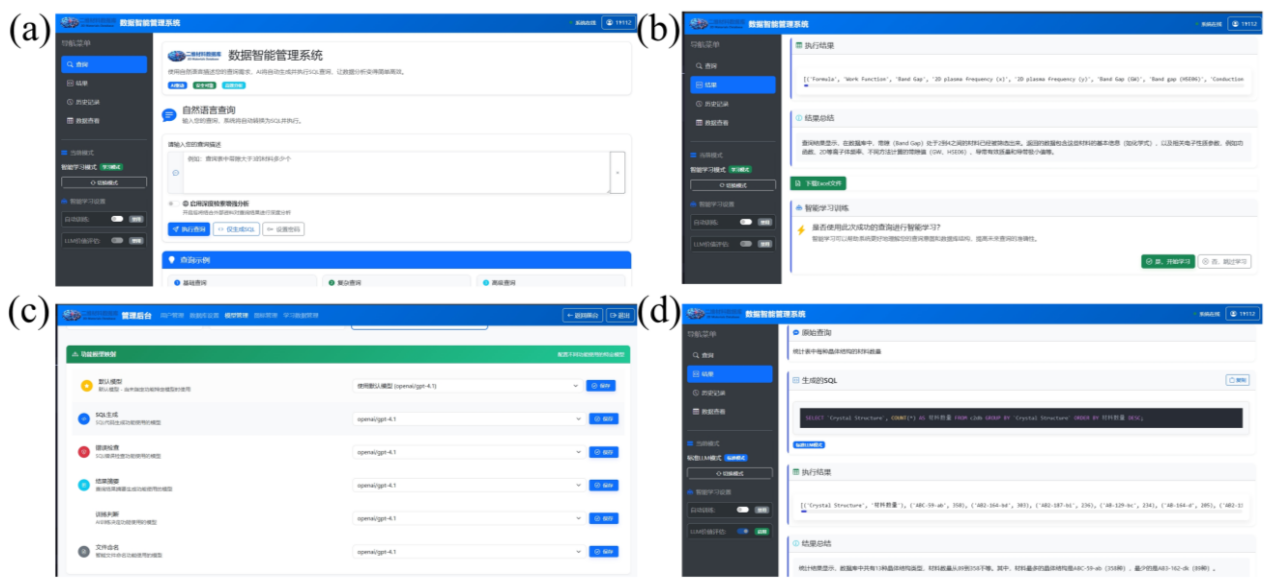

**Figure S1.** Overview of the data-intelligence management system and its functions: (a) overall system interface; (b) query results in intelligent-learning mode; (c) model management interface; (d) query results in standard mode.

Part C — Tables

1. Material Performance Sample:

| doi_or_title | material_name | parameter | value |
|---|---|---|---|
| Chem. Mater. 2022, 34, 2238−2248 | PbPc | density | 1.91 g/cm³ |
| Chem. Mater. 2022, 34, 2238−2248 | lead phthalocyanine (PbPc) | adsorption energy | -76.8 kcal/mol |
| Chem. Mater. 2022, 34, 2238−2248 | lead phthalocyanine (PbPc) | adsorption energy | -47.6 kcal/mol |
| Chem. Mater. 2022, 34, 2238−2248 | lead phthalocyanine (PbPc) | unit cell parameters | a = 52 ± 1 Å, b = 14 ± 1 Å, α = 88 ± 1° |
| Chem. Mater. 2022, 34, 2238−2248 | lead phthalocyanine (PbPc) | unit cell parameters | a = b = 14 ± 1 Å, α = 88 ± 1° |
| Chem. Mater. 2022, 34, 2238−2248 | lead phthalocyanine | gas sensitivity | NO2 sensor measurements |
| TiO2 nanoparticles-functionalized two-dimensional WO3 for highperformance supercapacitors developed by facile two-step ALD process | TiO2NP-WO3 | specific capacitance improvement | 1.5 times higher than pure 2D WO3 electrode |
| TiO2 nanoparticles-functionalized two-dimensional WO3 for highperformance supercapacitors developed by facile two-step ALD process | pure 2D WO3 electrode | specific capacitance | higher than most reported WO3 nanostructures |
| TiO2 nanoparticles-functionalized two-dimensional WO3 for highperformance supercapacitors developed by facile two-step ALD process | 2D TiO2NP-WO3 electrode | specific capacitance | 342.5, 327.1, 305.2, 290.6, and 285.3 F g−1 at current densities of 1.5, 3, 6, 15, and 30 A g−1 |
| TiO2 nanoparticles-functionalized two-dimensional WO3 for highperformance supercapacitors developed by facile two-step ALD process | 2D TiO2NP-WO3 electrode | capacitance retention | 94.7% after 2000 continuous charge-discharge cycles |
| TiO2 nanoparticles-functionalized two-dimensional WO3 for highperformance supercapacitors developed by facile two-step ALD process | pure 2D WO3 electrode | specific capacitance | 226.1, 208.5, 197.9, 194.6, and 190.3 F g−1 at current densities of 1.5, 3, 6, 15, and 30 A g−1 |
| TiO2 nanoparticles-functionalized two-dimensional WO3 for highperformance supercapacitors developed by facile two-step ALD process | pure 2D WO3 electrode | capacitance retention | 92.5% after 2000 continuous charge-discharge cycles |
| TiO2 nanoparticles-functionalized two-dimensional WO3 for highperformance supercapacitors developed by facile two-step ALD process | 2D WO3 | voltage range | 0–0.8 V |
| TiO2 nanoparticles-functionalized two-dimensional WO3 for highperformance supercapacitors developed by facile two-step ALD process | 2D WO3 | scan rate | 1–50 mV s−1 |
| TiO2 nanoparticles-functionalized two-dimensional WO3 for highperformance supercapacitors developed by facile two-step ALD process | 2D WO3 | current density | 1.5–30 A g−1 |
| TiO2 nanoparticles-functionalized two-dimensional WO3 for highperformance supercapacitors developed by facile two-step ALD process | 2D WO3 | cycling stability | measured at 6 A g−1 over 2000 cycles |
| TiO2 nanoparticles-functionalized two-dimensional WO3 for highperformance | 2D WO3 | thickness | 5.905 ± 0.325 nm |
| supercapacitors developed by facile two-step ALD process | | | |
| TiO2 nanoparticles-functionalized two-dimensional WO3 for highperformance supercapacitors developed by facile two-step ALD process | 2D TiO2NP-WO3 | voltage range | 0–0.8 V |
| TiO2 nanoparticles-functionalized two-dimensional WO3 for highperformance supercapacitors developed by facile two-step ALD process | 2D TiO2NP-WO3 | scan rate | 1–50 mV s−1 |
| TiO2 nanoparticles-functionalized two-dimensional WO3 for highperformance supercapacitors developed by facile two-step ALD process | 2D TiO2NP-WO3 | current density | 1.5–30 A g−1 |
| TiO2 nanoparticles-functionalized two-dimensional WO3 for highperformance supercapacitors developed by facile two-step ALD process | 2D TiO2NP-WO3 | molar amount of TiO2 | 10.76 mol% |
| TiO2 nanoparticles-functionalized two-dimensional WO3 for highperformance supercapacitors developed by facile two-step ALD process | 2D TiO2NP-WO3 | thickness | 1.587 ± 0.013 nm |
| TiO2 nanoparticles-functionalized two-dimensional WO3 for highperformance supercapacitors developed by facile two-step ALD process | 2D TiO2 | thickness | 1.587 ± 0.013 nm |
| TiO2 nanoparticles-functionalized two-dimensional WO3 for highperformance supercapacitors developed by facile two-step ALD process | 2D TiO2 | average growth rate | 0.71015 Å per cycle |
| TiO2 nanoparticles-functionalized two-dimensional WO3 for highperformance supercapacitors developed by facile two-step ALD process | 2D TiO2 | number of ALD cycles | 20 |
| Photonic enhanced flexible photothermoelectric generator using MXene coated fullerenes | PFM-100 film | Seebeck coefficient | 19.43 µV/K |
| Photonic enhanced flexible photothermoelectric generator using MXene coated fullerenes | PFM-100 film | Output voltage | 210 µV |
| Photonic enhanced flexible photothermoelectric generator using MXene coated fullerenes | PFM-100 film | Output voltage | 169 µV |
| Photonic enhanced flexible photothermoelectric generator using MXene coated fullerenes | PFM-100 film | Output voltage | 317 µV |
| Photonic enhanced flexible photothermoelectric generator using MXene coated fullerenes | PFM-100 film | Output voltage | 349 µV |
| Photonic enhanced flexible photothermoelectric generator using MXene coated fullerenes | PFM-100 film | PT output voltage | 2.02 mV |
| Photonic enhanced flexible photothermoelectric generator using MXene coated fullerenes | PFM-100 film | PT output voltage | 1.44 mV |
| Photonic enhanced flexible photothermoelectric generator using MXene coated fullerenes | PFM-100 film | Output voltage | 1.04 mV |
| Photonic enhanced flexible photothermoelectric generator using MXene coated fullerenes | PFM-100 film | Output voltage | 1.59 mV |
| Photonic enhanced flexible photothermoelectric generator using MXene coated fullerenes | PFM-100 film | Output voltage | 2.15 mV |
| Photonic enhanced flexible photothermoelectric generator using MXene coated fullerenes | PFM-100 film | Output voltage | 1.8 mV |
| Photonic enhanced flexible photothermoelectric generator using MXene coated fullerenes | PFM-100 film | VTE | 165.29 µV |
| Photonic enhanced flexible photothermoelectric generator using MXene coated fullerenes | PFM-100 film | VTE | 197.29 µV |
| Photonic enhanced flexible photothermoelectric generator using MXene coated fullerenes | PFM-100 | Seebeck coefficient | 18.13 µV/K |
| Photonic enhanced flexible photothermoelectric generator using MXene coated fullerenes | PFM-100 | output voltage at 15 mW | 1.04 mV |
| Photonic enhanced flexible photothermoelectric generator using MXene coated fullerenes | PFM-100 | output voltage at 20 mW | 1.59 mV |
| Photonic enhanced flexible photothermoelectric generator using MXene coated fullerenes | PFM-100 | output voltage at 25 mW | 2.15 mV |
| Photonic enhanced flexible photothermoelectric generator using MXene coated fullerenes | PFM-150 | Seebeck coefficient | 19.57 μV/K |
| Direct evidence of a charge depletion region at the interface of Van der Waals monolayers and dielectric oxides: The case of superconducting FeSe/STO | FeSe monolayer | distance from STO surface | 0.43 nm |
| Direct evidence of a charge depletion region at the interface of Van der Waals monolayers and dielectric oxides: The case of superconducting FeSe/STO | FeSe monolayer | depletion layer thickness | $6.5 \pm 1$ nm |
| Direct evidence of a charge depletion region at the interface of Van der Waals monolayers and dielectric oxides: The case of superconducting FeSe/STO | FeSe ML | superconducting gap | $11 \pm 3$ meV |
| Direct evidence of a charge depletion region at the interface of Van der Waals monolayers and dielectric oxides: The case of superconducting FeSe/STO | Nb-STO | plasma frequency | 83 meV |
| Direct evidence of a charge depletion region at the interface of Van der Waals monolayers and dielectric oxides: The case of superconducting FeSe/STO | Nb-STO | carrier concentration | $1.18 \times 10^{26}$ m$^{-3}$ |
| Direct evidence of a charge depletion region at the interface of Van der Waals monolayers and dielectric oxides: The case of superconducting FeSe/STO | FeSe | plasma frequency | 334 meV |
| Direct evidence of a charge depletion region at the interface of Van der Waals monolayers and dielectric oxides: The case of superconducting FeSe/STO | FeSe | effective mass | 3 |
| Direct evidence of a charge depletion region at the interface of Van der Waals monolayers and dielectric oxides: The case of superconducting FeSe/STO | FeSe | carrier density | 0.12 e$^-$/Fe |
| Direct evidence of a charge depletion region at the interface of Van der Waals monolayers and dielectric oxides: The case of superconducting FeSe/STO | Nb-STO | plasma frequency | 83 meV |
| Direct evidence of a charge depletion region at the interface of Van der Waals monolayers and dielectric oxides: The case of superconducting FeSe/STO | Nb-STO | plasmon linewidth (γpl) | 75 meV |
| Direct evidence of a charge depletion region at the interface of Van der Waals monolayers and dielectric oxides: The case of superconducting FeSe/STO | Nb-STO | plasmon linewidth ($\gamma_0$) | 5 meV |
| Direct evidence of a charge depletion region at the interface of Van der Waals monolayers and dielectric oxides: The case of superconducting FeSe/STO | STO | background dielectric constant | 5.9 |
| Prediction of synthesis of ternary-layered double transition metal MAX phases and the possibility of their exfoliation for formation of 2D MXenes | Mo2TiAlC2 | relative formation energy ($\Delta H$) | -0.082 eV |
| Prediction of synthesis of ternary-layered double transition metal MAX phases and the possibility of their exfoliation for formation of 2D MXenes | Cr2TiAlC2 | relative formation energy ($\Delta H$) | -0.031 eV |
| Prediction of synthesis of ternary-layered double transition metal MAX phases and the possibility of their exfoliation for formation of 2D MXenes | Mo2ScAlC2 | relative formation energy ($\Delta H$) | -0.037 eV |
| Prediction of synthesis of ternary-layered double transition metal MAX phases and the possibility of their exfoliation for formation of 2D MXenes | Mo2ScAlC2 | exfoliation energy (Eexf) | 0.202 eV/$Å^2$ |
| Prediction of synthesis of ternary-layered double transition metal MAX phases and the possibility of their exfoliation for formation of 2D MXenes | Mo2TiAlC2 | exfoliation energy (Eexf) | 0.201 eV/$Å^2$ |
| Prediction of synthesis of ternary-layered double transition metal MAX phases and the possibility of their exfoliation for formation of 2D MXenes | Cr2TiAlC2 | exfoliation energy (Eexf) | 0.206 eV/$Å^2$ |
| Optical properties of thin layers of rhenium diselenide (ReSe2) with thickness ranging from mono- (1 ML) to nona-layer (9 MLs) are demonstrated. | ReSe2 | in-plane force constant (Ks) | $18.9 \times 10^{19}$ N/$m^3$ |
| Optical properties of thin layers of rhenium diselenide (ReSe2) with thickness ranging from mono- (1 ML) to nona-layer (9 MLs) are demonstrated. | ReSe2 | out-of-plane force constant (Kb) | $69.6 \times 10^{19}$ N/$m^3$ |
| Optical properties of thin layers of rhenium diselenide (ReSe2) with thickness ranging from mono- (1 ML) to nona-layer (9 MLs) are demonstrated. | ReSe2 | blueshift of emission lines | about 120 meV |
| Optical properties of thin layers of rhenium diselenide (ReSe2) with thickness ranging from mono- (1 ML) to nona-layer (9 MLs) are demonstrated. | ReSe2 | energy separation between X1 and X2 peaks | about 55 meV for monolayer to almost 20 meV for 9 MLs |
| Optical properties of thin layers of rhenium diselenide (ReSe2) with thickness ranging from mono- (1 ML) to nona-layer (9 MLs) are demonstrated. | ReSe2 | emission energy for X1 peak | about 1.51 eV for 1 ML to ~1.39 eV for 9 MLs |
| Optical properties of thin layers of rhenium diselenide (ReSe2) with thickness ranging from mono- (1 ML) to nona-layer (9 MLs) are demonstrated. | ReSe2 | energy difference between two phonon modes | around 12 cm−1 for 1 ML to almost 6 cm−1 in 7-9 MLs |
| Optical properties of thin layers of rhenium diselenide (ReSe2) with thickness ranging from mono- (1 ML) to nona-layer (9 MLs) are demonstrated. | ReSe2 | energy difference between peaks 1 and 2 | 12 cm−1 to almost 6 cm−1 |
| Optical properties of thin layers of rhenium diselenide (ReSe2) with thickness ranging from mono- (1 ML) to nona-layer (9 MLs) are demonstrated. | ReSe2 | blueshift of phonon mode | about 120 meV |
| Optical properties of thin layers of rhenium diselenide (ReSe2) with thickness ranging from mono- (1 ML) to nona-layer (9 MLs) are demonstrated. | ReSe2 | blueshift of phonon mode | almost 4 cm−1 |
| Optical properties of thin layers of rhenium diselenide (ReSe2) with thickness ranging from mono- (1 ML) to nona-layer (9 MLs) are demonstrated. | ReSe2 | energy difference between peaks 1 and 2 | almost 12 cm−1 for a monolayer |
| Room temperature synthesis of freestanding 2D Mn3O4 nanostructures with enriched electrochemical properties for supercapacitor application | Mn3O4 | specific capacitance | 537 F/g (at 2 A g-1) |
| Room temperature synthesis of freestanding 2D Mn3O4 nanostructures with enriched electrochemical properties for supercapacitor application | Mn3O4 | specific capacitance retention | 93% (up to 5000 cycles) |
| Room temperature synthesis of freestanding 2D Mn3O4 nanostructures with enriched electrochemical properties for supercapacitor application | 2D Mn3O4 nanoplatelets | specific capacitance | 537 F/g at 2 A g-1 |
| Room temperature synthesis of freestanding 2D Mn3O4 nanostructures with enriched electrochemical properties for supercapacitor application | 2D Mn3O4 nanoplatelets | cyclic stability | 93% retention up to 5000 cycles |
| Room temperature synthesis of freestanding 2D Mn3O4 nanostructures with enriched electrochemical properties for supercapacitor application | 2D Mn3O4 nanoplatelets | specific capacitance | 236 F/g at 5 mV/s scan rate |
| Room temperature synthesis of freestanding 2D Mn3O4 nanostructures with enriched electrochemical properties for supercapacitor application | 2D Mn3O4 nanoplatelets | equivalent series resistance (Rs) | 5.28 Ohm |
| Room temperature synthesis of freestanding 2D Mn3O4 nanostructures with enriched electrochemical properties for supercapacitor application | 2D Mn3O4 nanoplatelets | charge transfer resistance (Rct) | 14.41 Ohm |
| Room temperature synthesis of freestanding 2D Mn3O4 nanostructures with enriched electrochemical properties for supercapacitor application | 2D Mn3O4 nanoplatelets | Warburg impedance (W1) | 45.7 Ohm |
| Room temperature synthesis of freestanding 2D Mn3O4 nanostructures with enriched electrochemical properties for supercapacitor application | 2D Mn3O4 nanoplatelets | specific capacitance | 537 F/g |
| Room temperature synthesis of freestanding 2D Mn3O4 nanostructures with enriched electrochemical properties for supercapacitor application | 2D Mn3O4 nanoplatelets | specific capacitance | 219 F/g |
| Room temperature synthesis of freestanding 2D Mn3O4 nanostructures with enriched electrochemical properties for supercapacitor application | 2D Mn3O4 nanoplatelets | specific capacitance | 137 F/g |
| Room temperature synthesis of freestanding 2D Mn3O4 nanostructures with enriched electrochemical properties for supercapacitor application | 2D Mn3O4 nanoplatelets | specific surface area | 39.06 m2/g |
| Room temperature synthesis of freestanding 2D Mn3O4 nanostructures with enriched electrochemical properties for supercapacitor application | 2D Mn3O4 nanoplatelets | average pore width | 2.76 nm |
| Room temperature synthesis of freestanding 2D Mn3O4 nanostructures with enriched electrochemical properties for supercapacitor application | 2D Mn3O4 nanoplatelets | crystallite size | 24.25 nm |
| Room temperature synthesis of freestanding 2D Mn3O4 nanostructures with enriched electrochemical properties for supercapacitor application | 2D Mn3O4 nanoplatelets | thickness | 3.5 nm |
| Room temperature synthesis of freestanding 2D Mn3O4 nanostructures with enriched electrochemical properties for supercapacitor application | 2D Mn3O4 nanoplatelets | thickness | 5.1 nm |
| Room temperature synthesis of freestanding 2D Mn3O4 nanostructures with enriched electrochemical properties for supercapacitor application | 2D Mn3O4 nanoplatelets | length | 100–200 nm |
| Room temperature synthesis of freestanding 2D Mn3O4 nanostructures with enriched electrochemical properties for supercapacitor application | 2D Mn3O4 nanoplatelets | width | 40–60 nm |
| Progress in the materials science of silicene | silicene | band gap at Dirac points (buckled silicene) | 1.55 meV |
| Progress in the materials science of silicene | silicene | band gap at Dirac points (buckled silicene with in-plane stress) | 2.90 meV |
| Progress in the materials science of silicene | silicene | band gap at Dirac points (planar silicene) | 0.07 meV |
| Progress in the materials science of silicene | silicene | Si–Si bond length in epitaxial silicene | 2.24 Å |
| Progress in the materials science of silicene | silicene | Si–Si bond length in bulk diamond-structured silicon | 2.35 Å |
| Progress in the materials science of silicene | silicene | Si–Si bond length in disilenes | 2.14–2.16 Å |
| Progress in the materials science of silicene | silicene | in-plane lattice constant of freestanding silicene | 3.855 Å |
| Progress in the materials science of silicene | silicene | in-plane lattice constant of Si(111) bilayer | 3.84 Å |

| doi_or_title | material_name | property_name | value |
|---|---|---|---|
| Progress in the materials science of silicene | silicene | band gap in epitaxial silicene nanoribbons | 0.5 eV |
| Progress in the materials science of silicene | silicene | energy of $\pi$ bands in epitaxial silicene nanoribbons | 0.6 eV below the Fermi level |
| Progress in the materials science of silicene | silicene | width of epitaxial silicene nanoribbons | 1.6 nm |
| Progress in the materials science of silicene | silicene | height of epitaxial silicene nanoribbons | 0.2 nm |
| Progress in the materials science of silicene | silicene on ZrB2(0001) | thermal stability | above 650 °C |
| Progress in the materials science of silicene | silicene on ZrC(111) | thermal stability | about 730 °C |
| Progress in the materials science of silicene | ZrB2(0001) | surface coverage with silicene | more than 99.5% |
| Progress in the materials science of silicene | ZrB2(0001) | lattice constant | about 3.65 Å |
| Progress in the materials science of silicene | ZrB2(0001) | compression relative to bulk Si(111) bilayer | 5% |
| Progress in the materials science of silicene | ZrB2(0001) | binding energy of Si 2p3/2 doublet | 600 ± 5 meV (spin–orbit splitting) |
| Progress in the materials science of silicene | ZrB2(0001) | binding energy shift of silicene 2p3/2 line | 450–710 meV towards lower binding energy |
| Progress in the materials science of silicene | ZrB2(0001) | energy gap observed by scanning tunneling spectroscopy | 350 meV, with center shifted 60 meV below the Fermi level |

## 2. Synthesis Methods Sample

| doi_or_title | material_name | method_name | method_details | reagents | conditions | equipment |
|---|---|---|---|---|---|---|

| Two dimensional (2D) materials such as graphene and transition metal dichalcogenides (TMDs) are promising for optical modulation, detection, and light emission since their material properties can be tuned on-demand via electrostatic doping | WS2 | Large-area synthesis | Large-area synthesis of high-quality and uniform monolayer WS2 on reusable Au foils. | Au foils | Not specified in the text | Not specified in the text |
|---|---|---|---|---|---|---|
| Two dimensional (2D) materials such as graphene and transition metal dichalcogenides (TMDs) are promising for optical modulation, detection, and light emission since their material properties can be tuned on-demand via electrostatic doping | WS2 | Synthesis of Large-Area WS2 monolayers | Synthesis of Large-Area WS2 monolayers with Exceptional Photoluminescence. | Not specified in the text | Not specified in the text | Not specified in the text |
| Two dimensional (2D) materials such as graphene and transition metal dichalcogenides (TMDs) are promising for optical modulation, detection, and light emission since their material properties can be tuned on-demand via electrostatic doping | MoS2 | Synthesis of MoS2 | Synthesis of MoS2 film for capacitive device experiments. | Not specified in the text | Not specified in the text | Not specified in the text |
| Two dimensional (2D) materials such as graphene and transition metal dichalcogenides (TMDs) are promising for optical modulation, detection, and light emission since their material properties can be tuned on-demand via electrostatic doping | monolayer WS2 | metal-organic chemical vapor deposition (MOCVD) | A 30 µm arc length of WS2 is grown using metal-organic chemical vapor deposition (MOCVD) and then patterned | WS2 (tungsten disulphide), ionic liquid [P14][FAP], Ti/Au contacts | Applied voltage of 2 V, room temperature, atmospheric exposure | Microring resonator, Ti/Au electrodes, ionic liquid [P14][FAP], COMSOL Multiphysics simulations |

| | | | onto the ring. The monolayer WS2 is electrically contacted with Ti/Au electrodes for gating. The ionic liquid [P14][FAP] is used for electrostatic doping, where a potential difference is applied between the electrode connected to WS2 and a nearby counter electrode, resulting in ion accumulation at the surface of WS2. The maximum electron doping density induced in the monolayer is about $7.2 \pm 0.8 \times 10^{13}$ cm^-2 with an applied voltage of 2 V. | | | |
| Two dimensional (2D) materials such as graphene and transition | monolay er WS2 | chemical vapor | CVD-grown films of WS2 are used for | WS2 (tungsten disulphide), | Voltage swing in the range | MZI (Mach Zehnder interferome |

| | | | | | |
|---|---|---|---|---|---|
| metal dichalcogenides (TMDs) are promising for optical modulation, detection, and light emission since their material properties can be tuned on-demand via electrostatic doping | deposition (CVD) | capacitive devices. These films are patterned on both arms of the MZI, followed by the deposition of metal electrodes. The ionic liquid [P14][FAP] is replaced with a stack of HfO2 (hafnia) and transparent conducting oxide ITO (indium tin oxide) to form the TMD-HfO2-ITO capacitor on the SiN waveguide. The charge neutrality point for the monolayer WS2 is at -4 V, corresponding to an initial electron doping of 1.5 ± 0.2 × 10^12 cm^-2, likely due to sulfur vacancies from CVD growth. | HfO2 (hafnia), ITO (indium tin oxide) | of -8 V to 9 V, room temperature | ter), HfO2-ITO capacitor, SU-8 photoresist for high index cladding, COMSOL Multiphysics simulations |

| | | | | | | |
|---|---|---|---|---|---|---|
| Two dimensional (2D) materials such as graphene and transition metal dichalcogenides (TMDs) are promising for optical modulation, detection, and light emission since their material properties can be tuned on-demand via electrostatic doping | WS2-HfO2-ITO capacitor | capacitive configuration | The WS2-HfO2-ITO capacitor is fabricated on each arm of the MZI. The monolayer WS2 is patterned on both arms of the MZI, followed by the deposition of metal electrodes. The ionic liquid [P14][FAP] is replaced with a stack of HfO2 (hafnia) and transparent conducting oxide ITO (indium tin oxide) to form the TMD-HfO2-ITO capacitor on the SiN waveguide. The charge neutrality point for the monolayer WS2 is at -4 V, corresponding to an initial electron doping of $1.5 \pm 0.2 \times 10^{12}$ cm$^{-2}$, likely | WS2 (tungsten disulphide), HfO2 (hafnia), ITO (indium tin oxide) | Voltage swing in the range of -8 V to 9 V, room temperature | MZI (Mach Zehnder interferometer), HfO2-ITO capacitor, SU-8 photoresist for high index cladding, COMSOL Multiphysics simulations |

| | | | due to sulfur vacancies from CVD growth. | | | |
|---|---|---|---|---|---|---|
| Rational and green synthesis of novel two-dimensional WS2/MoS2 heterojunction via direct exfoliation in ethanol-water targeting advanced visible-light-responsive photocatalytic performance | few-layer WS2 | Direct exfoliation in ethanol-water mixed solution | Commercial WS2 powders were dispersed in ethanol-water mixed solution to form a suspension with an initial concentration of 8 mg/mL. The suspension was then sonicated at 150 W for 8 h. After sonication, the dispersion underwent centrifugation at 3344 rpm for 30 min, and the supernatants containing few-layered WS2 were collected for further use. | WS2 powders, ethanol-water mixed solution | Sonication at 150 W for 8 h, centrifugation at 3344 rpm for 30 min | Ultrasonic bath, centrifuge |
| Rational and green synthesis of novel two-dimensional WS2/MoS2 heterojunction via direct exfoliation in ethanol-water targeting advanced visible-light-responsive photocatalytic performance | WS2/MoS2 heterostructure | In-situ hydrothermal reaction | Ammonium molybdate tetrahydrate (H32Mo7N6O28, 0.6 g) and thiourea (CN2H4S, 4.8 g) were dissolved | Ammonium molybdate tetrahydrate (0.6 g), thiourea (4.8 g), WS2 nanosheet suspension | Hydrothermal reaction at 200 °C for 24 h, vacuum drying at 60 °C overnight | Teflon-lined stainless-steel autoclave, vacuum oven, filtration system |

| | | | uniformly in 60 mL of WS2 nanosheet suspension (prepared in ethanol-water mixed solution) and stirred for 60 min. The mixture was sonicated for another 30 min at 28 kHz. It was then transferred into a Teflon-lined stainless-steel autoclave and reacted at 200 °C for 24 h. After cooling to room temperature, the black precipitates were collected by filtration, washed with DI water and ethanol, and dried in a vacuum oven at 60 °C overnight. | | | |
|---|---|---|---|---|---|---|
| Rational and green synthesis of novel two-dimensional WS2/MoS2 heterojunction via direct | WS2 nanosheets | Liquid exfoliation (LPE) | The starting material is commercialized WS2 | WS2 powder (Aladdin-reagent | Ethanol/water volume ratio of | Ultrasonic bath, centrifuge, UV–vis |

| | | | | |
|---|---|---|---|---|
| exfoliation in ethanol-water targeting advanced visible-light-responsive photocatalytic performance | powder (Aladdin-reagent Inc.). The powders are added into an ethanol-water mixture and subjected to sonication and centrifugation to form WS2 suspensions. By varying the volume ratio of ethanol and water, a series of suspensions containing uniformly-distributed WS2 nanosheets can be attained. The optimal condition is found to be an ethanol/water volume ratio of 7/3, which leads to the highest concentration of exfoliated WS2 (0.2 mg mL-1) and a dark greenish colour. This mild and green | Inc.), ethanol, water | 7/3 (v/v), sonication, centrifugation | spectrometer, photoluminescence (PL) spectrometer, SEM, TEM, AFM, Raman spectrometer |

| | | | exfoliation process produces uniform and stable WS2 inks without further purification. | | | |
|---|---|---|---|---|---|---|
| Rational and green synthesis of novel two-dimensional WS2/MoS2 heterojunction via direct exfoliation in ethanol-water targeting advanced visible-light-responsive photocatalytic performance | WS2/MoS2 heterojunctions | Hydrothermal synthesis | WS2/MoS2 heterojunctions are prepared using a hydrothermal route. LPE-derived WS2 nanosheets serve as 2D growth substrates. Ammonium molybdate tetrahydrate (H32Mo7N6O28, 0.6 g) and thiourea (CN2H4S, 4.8 g) are dissolved uniformly in 60 mL of WS2 nanosheet suspensions. The mixture undergoes a hydrothermal reaction to produce WS2/MoS2 heterojunctions via self-assembly. Different | Ammonium molybdate tetrahydrate (H32Mo7N6O28, 0.6 g), thiourea (CN2H4S, 4.8 g), WS2 nanosheet suspensions | Hydrothermal reaction, volume ratios of WS2 suspension/water (S/W): 3/7, 1/1, 7/3, and 4/1 | Hydrothermal reactor, SEM, TEM, EDS, XRD, XPS |

<table>
<tr><td></td><td></td><td></td><td>concentrations of WS2 suspensions are attempted (volume ratios of WS2 suspension/water: 3/7, 1/1, 7/3, and 4/1). The optimal condition is found to be an ethanol/water volume ratio of 7/3, which leads to the formation of blossoming flower-like structures instead of tightly-aggregated spheres due to the 2D WS2 template effect.</td><td></td><td></td><td></td></tr>
<tr><td>Protected giant magnetic anisotropy in two-dimensional materials: Transition-metal adatoms on defected tungsten disulfide monolayer</td><td>WS2</td><td>Defect-induced synthesis</td><td>The synthesis involves the creation of sulfur vacancies (Vs) in WS2 monolayers. These Vs can be either naturally formed during fabrication or deliberately produced in post</td><td>TM atoms (transition metals), WS2 monolayer with sulfur vacancies (Vs), H2S, S6, S8 (as common sulfur sources for WS2 preparation)</td><td>Low temperature, low concentration of TM atoms, presence of sulfur vacancies in WS2</td><td>Not specified</td></tr>
</table>

processes. The TM atoms are deposited onto the Vs sites, where they are strongly anchored due to high binding energies. The deposition is suggested to be done at low temperature and with a low concentration of TM atoms to prevent clustering. The surface remains reasonably flat with TM atoms slightly sticking out above the sulfur plane by a height of 0.15–0.40 Å. A van der Waals layer such as graphene or hex-BN can be added on top to protect the magnetic units from environmental interactions.

| | | | | | | |
|---|---|---|---|---|---|---|
| Protected giant magnetic anisotropy in two-dimensional materials: Transition-metal adatoms on defected tungsten disulfide monolayer | Os@D-WS2 | Defect-induced synthesis with Os atoms | Os atoms are deposited onto the sulfur vacancy (Vs) sites of D-WS2. The Os atoms are strongly anchored due to high binding energies. The deposition is done at low temperature and with a low concentration of Os atoms to prevent clustering. The Os atoms slightly protrude above the sulfur plane by a height of 0.15–0.40 Å. The resulting structure has a magnetic anisotropy energy (MAE) of 45 meV, which is sufficient to overcome thermal fluctuations for practical applications. | Os atoms, WS2 monolayer with sulfur vacancies (Vs) | Low temperature, low concentration of Os atoms, presence of sulfur vacancies in WS2 | Not specified |
| Protected giant magnetic anisotropy in two- | G/Os@D-WS2 | Graphene-adsorbed | A 5×5 graphene | Os atoms, WS2 | Fully relaxed | Not specified |

| dimensional materials: Transition-metal adatoms on defected tungsten disulfide monolayer | Os@D-WS2 synthesis | supercell is deposited on a 4×4 Os@D-WS2 supercell. The atomic structure is fully relaxed with the inclusion of the van der Waals correction term. The graphene layer aligns its C–C bridge site with the Os atom underneath, and the inter-planar distance between C and S is about 3.3 Å. The Os atoms are pulled out by 0.11 Å by the graphene layer. The resulting structure has a slightly reduced MAE of 38 meV but still remains among the largest reported. | monolayer with sulfur vacancies (Vs), graphene | atomic structure with van der Waals correction term included |

| https://doi.org/10.1016/j.ssc.2019.02.001 | T4,4,4-graphyne | First-principles calculations | The synthesis method involves designing a new structurally stable 2D carbon network composed of sp and sp2 hybridized atoms using first-principles calculations. The calculations are performed within the framework of density functional theory (DFT) using the Vienna ab initio simulation package (VASP). The electron exchange-correlation functional is treated using the generalized gradient approximation (GGA) with the Perdew-Burke-Ernzerhof (PBE) | Carbon atoms (sp and sp2 hybridized). | Energy cutoff of 520 eV, energy precision of 10^-5 eV, 15 × 15 × 1 Γ-centered Monkhorst-Pack grid for Brillouin zone sampling, maximum force on each atom less than 0.01 eV/Å during optimization, vacuum space of at least 17 Å in the perpendicular direction. | Vienna ab initio simulation package (VASP), PHONOPY code for phonon calculations. |

parameterization. An energy cutoff of 520 eV for the plane wave basis with an energy precision of $10^{-5}$ eV is used. The first Brillouin zone is sampled with a $15 \times 15 \times 1$ $\Gamma$-centered Monkhorst-Pack grid. During geometry optimization, both atomic positions and lattice vectors are fully optimized using the conjugate gradient scheme until the maximum force on each atom is less than 0.01 eV/Å. A large vacuum space of at least 17 Å is applied in the perpendicular direction of the 2D layer to avoid image interactions

| | | | from the periodic boundary condition. | | | |
|---|---|---|---|---|---|---|
| Evidence for Strain-Induced Local Conductance Modulations in Single-Layer Graphene on SiO2 | single-layer graphene on SiO2 | mechanical exfoliation | Graphene samples were manufactured via mechanical exfoliation. An optical microscope was used to identify the location of single-layer graphene, and photolithographic processes were employed to attach gold electrical contacts to the graphene. Prior to STM measurements, the sample was annealed in a high oxygen environment at temperatures of 400 oC for 15 minutes to remove photoresist that remained from the photolithography processes. | Photoresist (amount not specified), gold (for electrical contacts) | Annealing in a high oxygen environment at 400 oC for 15 minutes | Optical microscope, photolithographic equipment, STM (scanning tunneling microscope) |
| Dimensional Thermal-Hydraulic Analysis for a Packed Bed Regenerator Used in a Reheating Furnace | aluminum oxide spheres (Al2O3) | Packed bed stuffing preparation | The packed bed is filled with aluminum oxide spheres (Al2O3) with a diameter of 13 mm. The flow length for a unit cell is 474 mm and the width is 10.2 mm. The physical properties of the packed stuffing, such as thermal conductivity and specific heat, are considered as functions of temperature and expressed as polynomials of temperature. | Aluminum oxide (Al2O3) | Operating temperature varies during furnace cycles; thermal conductivity and specific heat are temperature-dependent and modeled as polynomials. | Commercial software (likely CFD-based) for simulation; no specific synthesis equipment is mentioned as the focus is on modeling and simulation. |
| | active MT liquid crystalline phase | hierarchical assembly | Active MT liquid crystalline phase is synthesized by first polymerizing microtubules (MTs) with GMPCPP to a concentration of 6 mg/ml. The filaments are left at room | microtubules (MTs), GMPCPP, biotin-labeled kinesin fragments (K401), tetrameric streptavidin, PEG (20kDa), phosphoenol pyruvate (PEP), | MTs polymerized at 6 mg/ml with GMPCPP; left at room temperature for 2 days to allow end-to-end annealing; ATP | conventional flow cell, fluorescence microscope, brightfield microscope |

| | | |
|---|---|---|
| temperature for 2 days to allow end-to-end annealing, resulting in an average length of 1.5 μm. Biotin-labeled kinesin fragments (401 amino acids of the N-terminal motor domain of D. melanogaster kinesin) are purified from E. Coli. Motor clusters are assembled by mixing biotin-labeled K401 with tetrameric streptavidin at a molar ratio of 1.7:1. A non-adsorbing polymer, PEG (20kDa), is added to induce attractive interactions through the depletion mechanism, forming extensile MT | pyruvate kinase/lactate dehydrogenase (PK/LDH), anti-bleaching agents, 3 μm beads, poly-L-lysine-poly(ethylene glycol) (PLL-PEG) block copolymers | concentration controlled via regenerating system; PEG induces depletion interactions; samples observed in a conventional flow cell with repulsive poly-acrylamide-coated surfaces |

bundles. The system is driven far from equilibrium by adding an ATP regenerating system composed of phosphoenol pyruvate (PEP) and pyruvate kinase/lactate dehydrogenase (PK/LDH), which maintains ATP concentration at a constant level. The active mixtures also include anti-bleaching agents. The resulting active MT networks are observed in a conventional flow cell with surfaces coated with a repulsive poly-acrylamide brush to prevent protein adsorption. The behavior

| | | | | | |
|---|---|---|---|---|---|
| | | of the active samples is monitored using fluorescence microscopy for MT visualization and brightfield microscopy for tracking 3 μm beads coated with poly-L-lysine-poly(ethylene glycol) (PLL-PEG) block copolymers to suppress sticking to MT bundles. | | | |
| active emulsion droplets | emulsification | Active emulsion droplets are created by dispersing BANs in an aqueous phase and forming a flat 2D oil-water interface stabilized with a surfactant. The surfactant used is PFPE-PEG-PFPE (E2K0660) in 3M HFE7500 oil. The | BANs (extensible MT bundles), PFPE-PEG-PFPE surfactant (E2K0660), 3M HFE7500 oil | aqueous phase emulsified in fluorinated oil; MT bundles adsorb onto droplet surfaces; droplets confined to quasi-2D chambers or spherical surfaces | flat 2D oil-water interface setup, emulsification system, quasi-2D chambers |

| | | | droplets are encapsulated in aqueous phase emulsified in fluorinated oil. When confined to quasi-2D chambers or spherical surfaces, the MT bundles adsorb onto the droplet surfaces, forming a dense liquid crystalline monolayer. The droplets exhibit autonomous motility when in frictional contact with a hard surface, driven by coherent internal flows generated by the active MT bundles. | | | |
| Facile assembly and electrochemical properties of a-Fe2O3@graphene aerogel composites as electrode materials for lithium ion batteries | Fe2O3@GA composites | Preparation of Fe2O3@GA composites | The working electrodes were prepared by mixing 80 wt% of active materials (Fe2O3), 10 wt% of super P as the conductive material, and | 80 wt% Fe2O3, 10 wt% super P, 10 wt% PVDF, NMP solvent | 120 °C, 12 h, vacuum oven | Scanning electron microscopy (SEM), Transmission electron microscopy (TEM), X-ray diffraction (XRD), N2 |

| | | | 10 wt% of polyvinyliden e fluoride (PVDF) as the binder in N-methylpyrroli dinone (NMP) solvent to form a slurry. The slurry was coated onto a copper foil and dried at 120 °C for 12 h in a vacuum oven. | | | adsorption-desorption isotherms |
|---|---|---|---|---|---|---|
| W2C/WS2 Alloy Nanoflowers as Anode Materials for Lithium-Ion Storage | W2C/WS 2_4h | hydrotherm al method | The W2C/WS2_4 h alloy nanoflowers (NFs) were successfully fabricated via a facile hydrothermal method at low temperature. The particle size of the NFs could be controlled between 200 nm and 1 µm. The obtained NFs exhibited high purity and well-defined hexagonal structures of W2C and WS2. The | W2C and WS2 | low temperatur e, hydrother mal conditions | hydrotherm al reactor |

| | | | control of reaction time for the preparation of W2C/WS2 electrodes is crucial to optimize the overall electrochemical properties, where the W2C/WS2_4 h alloy is the best electrode. | | | |
| --- | --- | --- | --- | --- | --- | --- |
| W2C/WS2 Alloy Nanoflowers as Anode Materials for Lithium-Ion Storage | WS2 NF | synthesis of WS2 nanoflowers | The as-prepared WS2 NFs showed lithiation at 0.5, 1.2, and 1.4 V, attributing to the reduction of WS2 to Li2S through multiple steps, including the formation of LixWS2 and Li2S. The WS2 NFs were used as anode materials for LIBs and showed charge and discharge capacities of 504.0 and 656.6 mAh | WS2 | 0.1–3.0 V voltage range | not specified |
| | | | g−1, respectively. | | | |
| Synthesis of PbCrO4 nanorods-Ti3C2Tx MXene composites: A sensitive photoelectrochemical sensor for the detection of cysteine in human blood serum | PbCrO4-MXene composite | Hydrothermal synthesis of PbCrO4 | PbCrO4 was prepared by hydrothermal method with slight modification. 5 mmol K2Cr2O7 was dissolved in 50 mL deionized water under continuous stirring to form solution A. 5 mmol Pb(CH3COO)2 was dissolved in 50 mL deionized water under continuous stirring to form solution B. Solutions A and B were mixed at a 1:2 volume ratio under continuous stirring. The mixture was transferred to a Teflon-lined 80-mL autoclave and kept at 150°C for 20 h. The product was centrifuged, washed with | K2Cr2O7 (5 mmol), Pb(CH3CO O)2 (5 mmol), deionized water (100 mL total), methanol (1 mL for composite preparation) | 150°C for 20 h (hydrother mal), 60°C vacuum drying, 300°C annealing for 2 h (composit e) | Teflon-lined autoclave, centrifuge, vacuum dryer, tube furnace |
| | | | deionized water, and dried under vacuum at 60°C overnight. | | | |
| Synthesis of $PbCrO_4$ nanorods-$Ti_3C_2T_x$ MXene composites: A sensitive photoelectrochemical sensor for the detection of cysteine in human blood serum | $PbCrO_4$-MXene composite | Composite preparation of $PbCrO_4$-MXene | 100 mg of $PbCrO_4$, 1 mL of monolayer MXene dispersion (5 mg/mL), and 1 mL methanol were dispersed in 50 mL deionized water. The mixture was sonicated for 1.5 h, centrifuged, and washed with water. The precipitate was dried overnight at 60°C under vacuum. The obtained powder was further annealed in a tube furnace at 300°C for 2 h. | $PbCrO_4$ (100 mg), MXene dispersion (5 mg/mL, 1 mL), methanol (1 mL), deionized water (50 mL) | 60°C vacuum drying, 300°C annealing for 2 h | Ultrasonic bath, centrifuge, vacuum dryer, tube furnace |
| AFM/XPS Analysis of the Growth and Architecture of Oriented Molecular Monolayer by Spin Cast Process and Its | $C_{32}H_{66}$ thin film | spin-coating | The $C_{32}H_{66}$ thin films were prepared by spin-coating at | $C_{32}H_{66}$ at concentrations of 0.13% w/v and 0.3% w/v | Spin-coating process with specific | AFM (Atomic Force Microscopy) for |
| Cross-Linking Induced by Hyperthermal Hydrogen | | | specific concentrations (e.g., 0.13% w/v, 0.3% w/v). The thickness of the thin film is controlled by the concentration and spin speed. The films were deposited on Si wafers and could be further treated with hexane washing and hyperthermal hydrogen bombardment for cross-linking. | | concentrations of $C_{32}H_{66}$; cross-linking was induced by hyperthermal hydrogen bombardment with fluence higher than $5 \times 10^{16}/cm^2$ and bombardment time up to 200 seconds. | imaging and thickness estimation; XPS (X-ray Photoelectron Spectroscopy) for surface chemical composition and thickness estimation; HHIC (Hyperthermal Hydrogen Ion Bombardment) equipment for cross-linking. |
| AFM/XPS Analysis of the Growth and Architecture of Oriented Molecular Monolayer by Spin Cast Process and Its Cross-Linking Induced by Hyperthermal Hydrogen | $C_{32}H_{66}$ nanofilm | Spin casting | $C_{32}H_{66}$ was dissolved into hexane under magnetic stirring, with its concentration varying from 0.05% w/v to 0.5% w/v. A 50 µL solution was casted on a Si wafer surface, spun at 5000 rpm until it transformed into a film, and the final | $C_{32}H_{66}$, hexane | Spin speed of 5000 rpm, drying at 60°C for 4 h | Spin-coating system |
| | | | C32H66 nanofilm was obtained after drying at 60°C for 4 h. | | | |
| AFM/XPS Analysis of the Growth and Architecture of Oriented Molecular Monolayer by Spin Cast Process and Its Cross-Linking Induced by Hyperthermal Hydrogen | C32H66 nanofilm | Hyperthermal hydrogen-induced cross-linking (HHIC) | Cross-linking experiments were conducted on a home-made electron cyclotron resonance (ECR)-enhanced microwave plasma reactor, which uses extraction ions to fill a space fully with H2 gas to collide and exchange its kinetic energy to hydrogen. This process may produce H with some certain kinetic energy and/or H2 having more kinetic energy. The energetic hydrogen breaks C-H bonds and then all C radicals will produce cross-linking. | Hydrogen (H2) | Kinetic energy of a few eV, via electron cyclotron resonance (ECR)-enhanced microwave plasma reactor | Home-made ECR-enhanced microwave plasma reactor |
| AFM/XPS Analysis of the Growth and Architecture of Oriented Molecular Monolayer by Spin Cast Process and Its Cross-Linking Induced by Hyperthermal Hydrogen | C32H66 | hypothermal hydrogen bombardment | The synthesis method involves inducing cross-linking in C32H66 thin-film surfaces through hyperthermal hydrogen bombardment. This process is described in the context of both fully and partially cross-linked surfaces, indicating that the bombardment is used to manipulate the degree of cross-linking in the material. | Hydrogen (H2) | Hyperthermal energy conditions are applied to the thin film surface. | Atomic force microscopy (AFM) is used for imaging and analysis. |
| AFM/XPS Analysis of the Growth and Architecture of Oriented Molecular Monolayer by Spin Cast Process and Its Cross-Linking Induced by Hyperthermal Hydrogen | dotriacontane | adsorption on SiO2 surface | Dotriacontane films are synthesized by adsorbing the alkane on a SiO2 surface. This is studied using high-resolution ellipsometry and synchrotron X-ray scattering techniques to | Dotriacontane (n-alkane) | Adsorption on a SiO2 surface under unspecified but controlled conditions. | Ellipsometry, synchrotron X-ray scattering, and molecular dynamics simulations. |

| | | | understand the structure and growth mode of the film. | | | |
|---|---|---|---|---|---|---|
| AFM/XPS Analysis of the Growth and Architecture of Oriented Molecular Monolayer by Spin Cast Process and Its Cross-Linking Induced by Hyperthermal Hydrogen | dotriacon tane | vapor-phase growth | In-situ X-ray reflectivity is used to study the growth of alkane films from the vapor phase. This method allows for the investigation of molecular orientation and film structure during deposition. | Alkane molecules | Vapor-phase growth conditions. | X-ray reflectivity equipment. |
| AFM/XPS Analysis of the Growth and Architecture of Oriented Molecular Monolayer by Spin Cast Process and Its Cross-Linking Induced by Hyperthermal Hydrogen | poly(trans-isoprene) | hypertherm al proton cross-linking | A cross-linking route for poly(trans-isoprene) involves the use of hyperthermal protons, which induce molecular cross-linking through unusual collision kinematics. This method allows for the synthesis of tailor-made molecular films with controlled | Hypertherm al protons | Collision-induced cross-linking under low chemical-and energy-loads. | Atomic force microscopy (AFM) and other surface analysis techniques. |

| | | | chemical properties. | | | |
|---|---|---|---|---|---|---|
| AFM/XPS Analysis of the Growth and Architecture of Oriented Molecular Monolayer by Spin Cast Process and Its Cross-Linking Induced by Hyperthermal Hydrogen | dodecane thiol on Au(111) | hypertherm al proton bombardme nt | Hyperthermal proton bombardment is applied to study its effects on self-assembled monolayers of dodecanethiol on Au(111). This method is used to investigate changes in the monolayer structure and properties. | Dodecaneth iol, hypertherm al protons | Bombard ment under controlled hyperther mal proton conditions . | Atomic force microscopy (AFM) and surface science techniques. |
| AFM/XPS Analysis of the Growth and Architecture of Oriented Molecular Monolayer by Spin Cast Process and Its Cross-Linking Induced by Hyperthermal Hydrogen | alkane-metal-system (C32H66 /In) | STM investigatio ns | Scanning tunneling microscopy (STM) is used to investigate the structure of the alkane-metal-system (C32H66/In). This method provides insights into the molecular arrangement and interactions at the interface. | C32H66, Indium (In) | Surface conditions suitable for STM imaging. | Scanning tunneling microscope (STM). |
| MXene-based promising nanomaterials for electrochemical energy storage | Ti3C2Tx | HF Etching | Ti3AlC2 MAX powder is added to hydrogen fluoride solution at | Ti3AlC2 MAX powder, hydrogen fluoride (HF) | Room temperatur e, pH ~6 after washing | Centrifuge, vacuum filtration |

room temperature. The reaction is exothermic, requiring gradual addition of MAX powder to avoid excessive bubble formation. The etching step produces a 2D-multilayered structure. The residue is washed with deionized water to remove AlF3 and sedimentary acids. After centrifugation and vacuum filtration at pH ~6, post-processing techniques can customize the final product. Surface termination with O, OH, and F groups occurs via fluoride and hydroxyl ions.

| | | | | | | |
|---|---|---|---|---|---|---|
| MXene-based promising nanomaterials for electrochemical energy storage | Ti3C2Tx | Fluoride Salt Etching (LiF/HCl) | LiF and HCl are mixed, followed by gradual addition of Ti3AlC2 MAX powder. The etching is heated at 40°C for 45 hours. By-products (LiCl, AlF3) are removed by washing. Water addition, centrifugation, and decanting increase pH. The final clay-type paste is rolled into a flexible MXene film using a roller mill. | LiF, HCl, Ti3AlC2 MAX powder | 40°C, 45 hours | Roller mill |
| MXene-based promising nanomaterials for electrochemical energy storage | Ti3C2Tx | NH4HF2 Etching | Ti3AlC2 MAX powder is treated with NH4HF2. The etching produces byproduct (NH4)3AlF6. Ammonium ions intercalate between Ti3C2Tx layers during the process. | NH4HF2, Ti3AlC2 MAX powder | Not explicitly stated | Not explicitly stated |

| | | | The final product forms thicker MXene films with larger inter-layer spaces due to intercalation. | | | |
|---|---|---|---|---|---|---|
| MXene-based promising nanomaterials for electrochemical energy storage | Ti3C2Tx | Alkali Treatment (NaOH) | Ti3AlC2 MAX powder is treated with strong alkaline compounds (NaOH). OH− ions oxidize Al into hydroxide, which dissolves in NaOH. By-products are removed via HF water solution and the Bayer process. High-temperature and high-alkali concentration conditions liquefy by-products to facilitate Al layer removal. | NaOH, Ti3AlC2 MAX powder | High temperature, high alkali concentration | Not explicitly stated |
| MXene-based promising nanomaterials for electrochemical energy storage | Ti3C2Tx | NaOH etching | Ti3C2Tx MXenes with 92 wt.% purity were fabricated by | 27.5 M NaOH | 270°C | Not specified |
| | | | treating Ti4AlN3 with 27.5 M NaOH at 270°C. | | | |
| MXene-based promising nanomaterials for electrochemical energy storage | Ti3C2Tx | Molten salt etching | Ti3C2Tx was synthesized by etching Ti3ZnC2 with ZnCl2 molten salt at 280°C (melting point) and 550°C. ZnCl2 ionizes into $Zn^{2+}$ and $ZnCl_4^{2-}$, oxidizing Al to $Al^{3+}$, forming AlCl3, and replacing Al with Zn. | ZnCl2 | 280°C (melting), 550°C (reaction), argon flow | Not specified |
| MXene-based promising nanomaterials for electrochemical energy storage | Ti3C2Tx | Electrochemical etching | Ti3C2Tx was produced via electrochemical etching of Ti3AlC2 in a tetramethylammonium hydroxide (TMA-OH) and NH4Cl aqueous electrolyte. $Cl^-$ corroded Al, breaking Ti-Al bonds, and NH4OH further etched the material. | TMA-OH, NH4Cl | Aqueous electrolyte, anode setup | Electrochemical cell |
| MXene-based promising nanomaterials for | Ti3C2Tx | Bottom-up sputtering and etching | Ti3AlC2 thin films were deposited on | HF or NH4HF2 | Not specified | Magnetron sputtering system |
| electrochemical energy storage | | | sapphire substrates via magnetron sputtering, followed by etching with HF or NH4HF2 to yield Ti3C2Tx. | | | |
| MXene-based promising nanomaterials for electrochemical energy storage | Ti3ZnC2 | Molten salt etching | Ti3ZnC2 was synthesized by etching with ZnCl2 molten salt at 280°C and 550°C, where Zn replaced Al in the MAX phase. | ZnCl2 | 280°C (melting), 550°C (reaction), argon flow | Not specified |
| MXene-based promising nanomaterials for electrochemical energy storage | Ti3C2Cl2 | Molten salt etching | Ti3C2Cl2 was produced by further etching Ti3ZnC2 with ZnCl2 molten salt. | ZnCl2 | 280°C (melting), 550°C (reaction), argon flow | Not specified |
| MXene-based promising nanomaterials for electrochemical energy storage | Mo2CTx | Thermal deposition and etching | Mo2CTx was synthesized by thermal deposition of Mo2GaC on a substrate, followed by HF etching to remove Ga layers. | HF | High temperature, HF etching | Thermal deposition system |
| MXene-based promising nanomaterials for electrochemical energy storage | Mo2TiC2Tx | Bottom-up synthesis | Mo2TiC2Tx was produced via bottom-up methods involving thermal deposition | Not specified | High temperature | Thermal deposition system |

| | | | and selective etching of MAX phases. | | | |
|---|---|---|---|---|---|---|
| 10.1080/02786826.2014.991438 | graphene nanoflowers (GNFs) | inductively annealing silicon-carbon (Si-C) nanoparticles | Graphene nanoflowers (GNFs) were synthesized by inductively annealing silicon-carbon (Si-C) nanoparticles at 2600℃, as described in Miettinen et al. (2011, 2014). The powder was produced using industrially scalable processes. | silicon-carbon (Si-C) nanoparticles | 2600℃ | inductively coupled annealing system |
| 10.1080/02786826.2014.991438 | graphene nanoflowers (GNFs) | ultrasonic spraying | Stable suspensions of GNFs were prepared in ethanol, DMF, and NMP. The suspensions were used to deposit micrometer-thick GNF films on glass substrates. The process involved an ultrasonic atomizer (Lechler, Type US 1) | ethanol, N,N-dimethylformamide (DMF), N-methyl-2-pyrrolidone (NMP) | 150℃ hot plate temperature, 9 L/min carrier gas flow rate, 10 cm spray distance, 200 mL/h suspension feeding rate, 7–14 min spraying time | ultrasonic atomizer (Lechler, Type US 1), syringe pump (Kd Scientific, KDS100), hot plate, plexiglass tube |

| | | | | | |
|---|---|---|---|---|---|
| | | | connected to a syringe pump (Kd Scientific, KDS100) and operated at 100 kHz. A carrier gas flow rate of 9 L/min was used to transport the droplets onto the substrate. The spray distance was 10 cm, and the suspension feeding rate was 200 mL/h. The hot plate temperature was set to 150°C (actual measured temperature 111°C). The spraying time varied between 7 min and 14 min, with some tests using 30 s for early deposition analysis. | | |
| 10.1080/02786826.2014.991438 | multi-layer graphene (MLG) flakes | inductively annealing silicon-carbon (Si-C) | Multi-layer graphene (MLG) flakes were synthesized | silicon-carbon (Si-C) nanoparticles | 2600°C | inductively coupled annealing system |
| | nanoparticles | | by inductively annealing silicon-carbon (Si-C) nanoparticles at 2600°C, as described in Miettinen et al. (2011, 2014). The powder was produced using industrially scalable processes. | | | |
| 10.1080/02786826.2014.991438 | multi-layer graphene (MLG) flakes | ultrasonic spraying | Stable suspensions of MLG flakes were prepared in ethanol, DMF, and NMP. The suspensions were used to deposit micrometer-thick MLG films on glass substrates. The process involved an ultrasonic atomizer (Lechler, Type US 1) connected to a syringe pump (Kd Scientific, KDS100) and operated at | ethanol, N,N-dimethylformamide (DMF), N-methyl-2-pyrrolidone (NMP) | 150°C hot plate temperature, 9 L/min carrier gas flow rate, 10 cm spray distance, 200 mL/h suspension feeding rate, 7–14 min spraying time | ultrasonic atomizer (Lechler, Type US 1), syringe pump (Kd Scientific, KDS100), hot plate, plexiglass tube |
| | | | 100 kHz. A carrier gas flow rate of 9 L/min was used to transport the droplets onto the substrate. The spray distance was 10 cm, and the suspension feeding rate was 200 mL/h. The hot plate temperature was set to 150°C (actual measured temperature 111°C). The spraying time varied between 7 min and 14 min, with some tests using 30 s for early deposition analysis. | | | |
| 10.1080/02786826.2014.991438 | graphene nanoflowers (GNFs) | ultrasonic spraying | GNF suspensions with a concentration of 0.05 wt% were ultrasonically sprayed onto glass for 7 min and 14 min. The | 0.05 wt% GNF suspension in DMF/ethanol solvent mixture | Spraying time of 7 min and 14 min, substrate temperature of approximately 111 C (hot plate | Ultrasonic sprayer |

| | | | films were prepared using a DMF/ethanol solvent mixture due to its lower boiling point compared to NMP. The process resulted in percolated films with substrate coverages of approximately 70% and 79% for 7-min and 14-min depositions, respectively. The films showed a rough surface due to the inherently irregular shape of the graphene nanomaterials. | | set to 150 C) | |
|---|---|---|---|---|---|---|
| 10.1080/02786826.2014.991438 | Multi-Layer Graphene (MLG) flakes | ultrasonic spraying | MLG suspensions with concentrations of 0.05 wt%, 0.1 wt%, and 0.2 wt% were ultrasonically sprayed for 7 min onto | 0.05 wt%, 0.1 wt%, and 0.2 wt% MLG suspension in DMF/ethanol solvent mixture | Spraying time of 7 min, substrate temperature of approximately 111 C (hot plate set to 150 C) | Ultrasonic sprayer |

| | | | glass substrates. The films were prepared using a DMF/ethanol solvent mixture. The morphology of MLG films on glass was investigated, and the films were found to be composed of compact agglomerates at higher concentrations. The process resulted in a substrate coverage of approximately 79% for 0.2 wt% MLG suspensions. The films showed a rough surface due to the irregular shape of the graphene nanomaterials. | | | |
|---|---|---|---|---|---|---|
| 10.1080/02786826.2014.991438 | Graphene Nanoplatelets (GNF) | Electrospray Deposition | Graphene nanoplatelets (GNF) are deposited using electrospray deposition. The process | Graphene nanoplatelets (GNF) in suspensions | Spraying time of 14 minutes, 0.05 wt% GNF suspensions | Electrospray deposition equipment |

| | | | involves suspensions of GNF in solvents and is applied to MA-glass slides. The spraying time is critical, with nearly 100% substrate coverage achieved at 14 minutes of spraying using 0.05 wt% GNF suspensions. The method inherently involves rotation of individual graphene layers with respect to each other, as revealed by Raman spectra. | | | |
| 10.1080/02786826.2014.991438 | Few-Layer Graphene Nanosheets | Synthesis in Ethanol-Soluble Dispersions | Few-layer graphene nanosheets are synthesized using a chemical route that allows them to be ethanol-soluble. This method is used to | Graphene precursors for exfoliation | Ethanol-soluble conditions | Chemical synthesis and dispersion equipment |
| | | | produce flexible and transparent conducting composite films. The process involves chemical exfoliation and dispersion techniques to achieve the desired solubility and film properties. | | | |
| 10.1080/02786826.2014.991438 | N-Doped Graphene Nanoplatelets | Chemical Doping and Exfoliation | N-doped graphene nanoplatelets are synthesized through a chemical doping process involving nitrogen incorporation. The method is used to create superior metal-free counter electrodes for organic dye-sensitized solar cells. The process includes exfoliation and doping steps to | Graphene precursors and nitrogen-doping agents | Conditions for exfoliation and nitrogen doping | Chemical synthesis and exfoliation equipment |

| | | | achieve the desired electronic properties. | | | |
|---|---|---|---|---|---|---|
| 10.1080/02786826.2014.991438 | Reduced Graphene Oxide Sheets | Covalent Assembly | Reduced graphene oxide sheets are synthesized via covalent assembly on a silicon substrate. This method involves chemical functionalization and assembly to produce the graphene sheets. The process is used to study tribology on silicon substrates. | Graphene oxide and functionalizing agents | Chemical conditions for covalent assembly | Chemical synthesis and coating equipment |
| 10.1080/02786826.2014.991438 | Silicon–Carbon Nanoceramics | Atmospheric Pressure Chemical Vapour Synthesis | Silicon–carbon nanoceramics are synthesized using atmospheric pressure chemical vapour synthesis from hexamethyldisilane in a high-temperature aerosol | Hexamethyl disilane | High-temperature aerosol reactor, atmospheric pressure | High-temperature aerosol reactor |

| | | | reactor. The method involves the decompositio n of hexamethyldi silane at high temperatures to form the nanoceramics . | | | |
|---|---|---|---|---|---|---|
| 10.1080/02786826.2014. 991438 | Graphene Oxide Films | Electrospra y Deposition and Post-Annealing | Graphene oxide films are synthesized using electrospray deposition followed by post-annealing. The electrospray process deposits the graphene oxide, and the post-annealing step enhances crystallinity and reduces defects. The sensitivity of the micro-Raman spectrum to the crystallite size of the films is studied. | Graphene oxide | Electrospr ay deposition parameter s and post-annealing conditions | Electrospra y deposition system and annealing furnace |
| 10.1080/02786826.2014. 991438 | Graphene -Based | Electrodyna mically | Graphene-based thin films are | Graphene nanosheets | Electrody namic | Electrospra y |

| | | Thin Films | Sprayed Deposition | synthesized using electrodynamically sprayed deposition. This method is used for device applications and involves the controlled deposition of graphene nanosheets to form thin films with specific properties. | | spraying conditions | deposition system |
|---|---|---|---|---|---|---|---|
| 10.1080/02786826.2014.991438 | | Nickel Oxide-Gadolinia Doped Ceria Composite Thin Films | Spray Pyrolysis | Nickel oxide-gadolinia doped ceria composite thin films are synthesized using spray pyrolysis. The method involves the Leidenfrost effect during the deposition process, which affects the film formation and properties. The process is used to create composite thin films with specific functionalities. | Nickel oxide and gadolinia doped ceria precursors | Spray pyrolysis conditions involving the Leidenfrost effect | Spray pyrolysis system |

| | | | | | | |
|---|---|---|---|---|---|---|
| 10.1080/02786826.2014.991438 | Zirconia Coatings | Electrostatic Spray Deposition (ESD) | Zirconia coatings are synthesized using electrostatic spray deposition (ESD). The method involves the application of an electric field to deposit zirconia particles onto a substrate. The initial stages of coating formation are studied to understand the process parameters and their effects on the final film properties. | Zirconia precursors | Electrostatic spray deposition parameters | Electrostatic spray deposition system |
| 10.1080/02786826.2014.991438 | Graphene Nanosheets | Chemical Route | Graphene nanosheets are synthesized via a chemical route suitable for device applications. The method involves chemical exfoliation and functionalizat | Graphene precursors and chemical exfoliation agents | Chemical exfoliation and functionalization conditions | Chemical synthesis and exfoliation equipment |
| | | | ion to produce nanosheets with specific properties for integration into devices. | | | |
| 10.1080/02786826.2014.991438 | Graphene Oxide Films | Low Temperature Reduction with Vanadium(III) | Graphene oxide films are synthesized and then reduced at low temperatures using vanadium(III). The reduction process is effective at low temperatures and is used to produce graphene films with specific electronic and structural properties. | Graphene oxide, vanadium(III) | Low-temperature reduction conditions | Reduction furnace or low-temperature reactor |
| 10.1038/s41598-020-68321-7 | WS2 | CVD synthesis | The CVD synthesis method is used to produce WS2 films, which favor small grains (~10 nm²), but larger grains up to approximatel | Not specified in the text | Not specified in the text | Not specified in the text |

| | | | y 100 nm² are also present. | | | |
|---|---|---|---|---|---|---|
| 10.1038/s41598-020-68321-7 | WS2 | ALD and gas conversion | 2 nm of WO3 is deposited using ALD onto 30 nm thick electron transparent Si3N4 TEM windows. The WO3 is then converted to WS2 by flowing H2S gas over the films at elevated temperature. | WO3, H2S gas | Elevated temperature | ALD system, electron transparent Si3N4 TEM windows |
| 10.3390/biomedicine9010080 | DOX/Pd @ZIF-8@PDA | self-assembled one-pot method | DOX, Pd, and ZIF-8 were conjugated in a self-assembled one-pot method, and then coated with PDA to improve biocompatibility. | DOX, Pd, ZIF-8, PDA | Not specified in the provided text | Not specified in the provided text |
| 10.3390/biomedicine9010080 | MoS2-PEG nanosheet | Preparation of MoS2-PEG nanosheet | The nanosheets were prepared by loading doxorubicin (DOX) onto MoS2-PEG. The nanosheets were analyzed for drug delivery and photothermal | Doxorubicin (DOX), MoS2, PEG | NIR irradiation | Not specified |

| | | | effects by NIR irradiation. | | | |
|---|---|---|---|---|---|---|
| 10.3390/biomedicine901 0080 | MnO2-PEG-cRGD/C e6 | Preparation of MnO2-PEG-cRGD/Ce6 | Ultrathin MnO2 nanosheets were prepared with polyethylene glycol-cyclic arginine-glycine-aspartic acid tripeptide (PEG-cRGD) and encapsulated chlorin e6 (Ce6). The nanosheets showed photothermal efficiency of 39% and pH-controlled Ce6 release under a 660 nm NIR laser. | MnO2, PEG-cRGD, chlorin e6 (Ce6) | pH-controlled release, 660 nm NIR laser | Not specified |
| 10.3390/biomedicine901 0080 | TiL4@B PQDs | Fabrication of TiL4@BPQ Ds | TiL4 was coordinated with black phosphorus quantum dots (BPQDs) to enhance their stability in aqueous dispersions. The TiL4@BPQD s were used for PA imaging of cancer cells | TiL4, black phosphorus quantum dots (BPQDs) | Aqueous dispersion | Not specified |

| | | | and xenografted tumors in mice. | | | |
|---|---|---|---|---|---|---|
| 10.3390/biomedicine9010080 | ICG-rGO nanocomposite | Fabrication of ICG-rGO nanocomposite | Indocyanine green (ICG) was conjugated to reduced graphene oxide (rGO) to create a nanocomposite for dual use in PA imaging and fluorescence imaging. The nanocomposite showed high PA/fluorescence signals and strong tumor accumulation. | Indocyanine green (ICG), reduced graphene oxide (rGO) | NIR regime | Not specified |
| 10.3390/biomedicine9010080 | AMNFs (Antimony nanoflakes) | Liquid phase exfoliation of AMNFs | AMNFs were fabricated via liquid phase exfoliation of bulk antimony (Sb). The nanoflakes exhibited unique PA properties and high PT conversion efficiency (42.36%). | Bulk antimony (Sb) | Liquid phase exfoliation | Not specified |
| 10.3390/biomedicine9010080 | Sb2Se3 nanosheets | Fabrication of Sb2Se3 nanosheets | Sb2Se3 nanosheets were fabricated | Antimony (Sb), Selenium (Se) | Liquid nitrogen pretreatment, freeze– |  |

| | | | using liquid nitrogen pretreatment and a freeze–thawing approach. The nanosheets had an extinction coefficient of 33.2 L/g·cm and a photothermal conversion efficiency of 30.78%. | | thawing approach | |
| 10.3390/biomedicine9010080 | Ti3C2 MXene | Exfoliation of Ti3C2 MXene | Ti3C2 MXene was exfoliated from layer-structured MAX-phase ceramics. The nanosheets exhibited a PA signal at an excitation wavelength of 808 nm in vitro and a significant increase in PA signal at a concentration of 15 mg/kg in tumor-bearing mice. | MAX-phase ceramics | Exfoliation from MAX-phase ceramics | Not specified |
| 10.3390/biomedicine9010080 | SP-MnOx/Ta4C3 composite nanosheets | Fabrication of SP-MnOx/Ta4C3 composite nanosheets | SP-modified MnOx/Ta4C3-based nanosheets were fabricated. The | MnOx, Ta4C3, SP (stabilizing agent) | Subcutaneous administration | Not specified |

| | | | nanosheets showed fast enhanced PA signals after subcutaneous administratio n in 4T1 tumor-bearing mice and exhibited contrast-enhanced properties in CT and MR. | | | |
|---|---|---|---|---|---|---|
| 10.3390/biomedicine901 0080 | PVP-Bi2Se3 nanoshee ts | Fabrication of PVP-Bi2Se3 nanosheets | PVP-encapsulated Bi2Se3 nanosheets were fabricated. The nanosheets had an extinction coefficient of 11.5 L/g·cm at 808 nm, a PT conversion efficiency of 34.6%, and excellent photoacoustic performance with a twofold increase after 5 h of injection. | Bi2Se3, PVP (polyvinyl pyrrolidone) | 808 nm excitation wavelengt h, 5 h post-injection | Not specified |
| 10.3390/biomedicine901 0080 | HA-MoS2 nanoshee ts | Fabrication of HA-MoS2 nanosheets | HA (hyaluronic acid) was conjugated to MoS2 | MoS2, HA (hyaluronic acid) | pH and GSH-induced intracellul ar | Not specified |
| | | | nanosheets to induce HA receptor-mediated endocytosis. The nanosheets were internalized via CD44, HARE, and LYVE-1 receptors, and showed a 1860-fold increase in PA signal in tumor-bearing mice. | | | disulfide cleavage |
| 10.3390/biomedicine9010080 | Pd nanosheets | Fabrication of Pd nanosheets | Pd nanosheets were fabricated as typical 2D nanomaterials with well-defined and controllable sizes. The nanosheets were used for PA-guided PTT due to their strong NIR absorption and excellent biocompatibility. | Palladium (Pd) | NIR absorption | Not specified |
| Multilayered evanescent wave absorption based fluoride fiber sensor with 2D material and amorphous silicon layers for enhanced sensitivity | MoS2 monolayer | Coating | A 5-layer sensor design is simulated with a MoS2 monolayer coated over a | MoS2 monolayer, $\alpha$-Si layer (10 nm thick), ZBLA core, | Near-infrared (NIR) spectral region, $\lambda$ = 0.954 μm, | Fiber optic sensor system with 5-layer configuration (ZBLA |

| and resolution in near infrared | 10 nm thick α-Si layer. The sensor is analyzed for its performance in the near-infrared (NIR) spectral region for detecting malignancy in liver tissues through variations in their refractive index (RI) values. The sensitivity and resolution are evaluated for different configurations, including varying the thickness of the α-Si layer and the cladding material. The dispersion relation of H2O is considered as a reference analyte, which leads to improved sensitivity and resolution compared to using N tissue | 19.3% GeO2 doped silica clad, H2O as reference analyte, HCC tissue as analyte | $\lambda = 1.236$ μm, $\lambda = 1.55$ μm | core – 19.3% GeO2 doped silica clad – MoS2 monolayer – α-Si layer – analyte) |
|---|---|---|---|---|

| | | | | | | |
|---|---|---|---|---|---|---|
| | | | as a reference. The maximum sensitivity values (Sr, Si) for H2O-HCC detection are reported as (225.5 mW/RIU, 1182.9 mW/RIU) at λ = 0.954 µm. | | | |
| Multilayered evanescent wave absorption based fluoride fiber sensor with 2D material and amorphous silicon layers for enhanced sensitivity and resolution in near infrared | 4-layer sensor design with 10 nm thick α-Si layer | Simulation | A 4-layer sensor design is simulated without a 2D material, consisting of a ZBLA core, 19.3% GeO2 doped silica clad (10 nm thick), and analyte. The sensitivity and resolution are compared with the 5-layer sensor design with MoS2 and α-Si layers. At λ = 1.236 µm, the sensitivity values (Sr, Si) are reported as (110.3 mw/RIU, 647.8 mw/RIU). | ZBLA core, 19.3% GeO2 doped silica clad (10 nm thick), analyte | Near-infrared (NIR) spectral region, λ = 1.236 µm | Fiber optic sensor system with 4-layer configuration (ZBLA core – 19.3% GeO2 doped silica clad – analyte) |

| Multilayered evanescent wave absorption based fluoride fiber sensor with 2D material and amorphous silicon layers for enhanced sensitivity and resolution in near infrared | 5-layer sensor design with 10 nm thick α-Si layer | Simulation with MoS2 and α-Si layers | A 5-layer sensor design is simulated with a ZBLA core, 19.3% GeO2 doped silica clad, a MoS2 monolayer, a 10 nm thick α-Si layer, and analyte. The sensor is analyzed for sensitivity and resolution in the NIR region. The sensitivity values (Sr, Si) at λ = 1.236 μm are (123.0 mw/RIU, 653.9 mw/RIU), and the resolution values (Rr, Ri) are (8.13 × 10−10 RIU, 1.53 × 10−10 RIU). The sensor design with 10 nm α-Si layer shows a 11.5% enhancement in Sr compared to the 4-layer sensor without α-Si. | ZBLA core, 19.3% GeO2 doped silica clad, MoS2 monolayer, α-Si layer (10 nm thick), analyte | Near-infrared (NIR) spectral region, λ = 1.236 μm | Fiber optic sensor system with 5-layer configuration (ZBLA core – 19.3% GeO2 doped silica clad – MoS2 monolayer – α-Si layer – analyte) |
|---|---|---|---|---|---|---|

| Multilayered evanescent wave absorption based fluoride fiber sensor with 2D material and amorphous silicon layers for enhanced sensitivity and resolution in near infrared | 5-layer sensor design with 10 nm thick α-Si layer and H2O as reference analyte | Simulation with H2O reference | A 5-layer sensor design is simulated with a ZBLA core, 19.3% GeO2 doped silica clad, a MoS2 monolayer, a 10 nm thick α-Si layer, and H2O as the reference analyte. The sensitivity and resolution are evaluated for both real and imaginary parts of HCC tissue RI. At $\lambda = 0.954 \mu m$, the sensitivity values (Sr, Si) are (225.5 mW/RIU, 1182.9 mW/RIU), and the resolution values (Rr, Ri) are (4.43 $\times$ 10−10 RIU, 8.85 $\times$ 10−11 RIU). This design is found to provide 1.8 times better sensitivity than the N-HCC | ZBLA core, 19.3% GeO2 doped silica clad, MoS2 monolayer, α-Si layer (10 nm thick), H2O as reference analyte, HCC tissue as analyte | Near-infrared (NIR) spectral region, $\lambda = 0.954 \mu m$ | Fiber optic sensor system with 5-layer configuration (ZBLA core – 19.3% GeO2 doped silica clad – MoS2 monolayer – α-Si layer – analyte) |

| | | | detection design. | | | |
|---|---|---|---|---|---|---|
| 10.1021/acsnano.2c0106 5 | MoS2 | High-pressure Raman and photoluminescence (PL) experiments | Monolayer MoS2 samples were prepared via chemical vapor deposition (CVD) onto SiO2/Si substrates and loaded into a diamond anvil cell (DAC) using the horseshoe method. A 2.33 eV excitation laser was used at 10 K to collect Raman and PL spectra. A 4:1 methanol/ethanol mixture was used as the pressure transmitting medium (PTM). Pressure was determined in situ using a low-temperature ruby fluorescence calibration. | Methanol, ethanol, 2.33 eV excitation laser | Temperature: 10 K, Pressure: 0−4.5 GPa, Atmosphere: Hydrostatic pressure using DAC | Diamond anvil cell (DAC), Raman spectrometer, photoluminescence setup, 2.33 eV laser |
| 10.1021/acsnano.2c0106 5 | WSe2 | High-pressure Raman and | Monolayer WSe2 samples were | Methanol, ethanol, 2.33 eV | Temperature: 10 K, Pressure: | Diamond anvil cell (DAC), |

| | | photolumin escence (PL) experiments | prepared via chemical vapor deposition (CVD) onto SiO2/Si substrates and loaded into a diamond anvil cell (DAC) using the horseshoe method. A 2.33 eV excitation laser was used at 10 K to collect Raman and PL spectra. A 4:1 methanol/etha nol mixture was used as the pressure transmitting medium (PTM). Pressure was determined in situ using a low-temperature ruby fluorescence calibration. | excitation laser | 0−4.5 GPa, Atmosphe re: Hydrostati c pressure using DAC | Raman spectromete r, photolumin escence setup, 2.33 eV laser |
|---|---|---|---|---|---|---|
| 10.1021/acsnano.2c0106 5 | MoS2 | metal−orga nic chemical vapor deposition (MOCVD) | Monolayer MoS2 flakes were synthesized on a 300 nm-thick SiO/Si wafer using a one-inch | molybdenu m hexacarbon yl (Mo(CO)6), diethyl sulfide ((C2H5)2S), | low pressure, growth temperatur e below 350 °C, growth time of 6 h | one-inch quartz tube furnace, bubbler system |

| | | | quartz tube furnace under low pressure. The precursors used were molybdenum hexacarbonyl (Mo(CO)6) and diethyl sulfide ((C2H5)2S). The synthesis was carried out with an Ar carrier gas flow rate of 100 sccm, 0.1 sccm for Mo(CO)6, and 1.0 sccm for (C2H5)2S. The growth temperature was below 350 °C, and the growth time was 6 hours. | Ar carrier gas | | |
|---|---|---|---|---|---|---|
| 10.1021/acsnano.2c01065 | WSe2 | metal−organic chemical vapor deposition (MOCVD) | Monolayer WSe2 flakes were synthesized using a one-inch quartz tube furnace under low pressure. The precursors used were tungsten hexacarbonyl (W(CO)6), | tungsten hexacarbonyl (W(CO)6), dimethyl selenide ((CH3)2Se), Ar carrier gas, H2 | low pressure, growth temperature of 420 °C, growth time of 5 h | one-inch quartz tube furnace, bubbler system |

| | | | dimethyl selenide ((CH3)2Se), and hydrogen (H2). The synthesis was carried out with an Ar carrier gas flow rate of 100 sccm, 1 sccm for H2, 0.3 sccm for W(CO)6, and 0.05 sccm for (CH3)2Se. The growth temperature was 420 °C, and the growth time was 5 hours. | | | |
|---|---|---|---|---|---|---|
| 10.1021/acsnano.2c01065 | monolayer MoS2 | chemical vapor deposition | The synthesis of high-performance monolayer molybdenum disulfide is achieved at low temperature using chemical vapor deposition (CVD). | molybdenum and sulfur precursors | low temperature | chemical vapor deposition system |
| 10.1021/acsnano.2c01065 | monolayer MoSe2 | chemical vapor deposition | Monolayer MoSe2 devices are fabricated by the pick-up of pre-patterned hBN, enabling | molybdenum and selenium precursors | not explicitly mentioned | chemical vapor deposition system, hBN (hexagonal boron nitride) for |

| | | | encapsulation and precise electronic transport studies. | | | encapsulation |
|---|---|---|---|---|---|---|
| 10.1021/acsnano.2c01065 | monolayer WSe2 | chemical vapor deposition | Monolayer WSe2 is studied in the context of interlayer excitons and their control through low-temperature photocurrent spectroscopy. | tungsten and selenium precursors | low temperature | chemical vapor deposition system |
| 10.1021/acsnano.2c01065 | few-layer graphene | compression | A hard, transparent, sp3-containing 2D phase is formed from few-layer graphene under compression. | few-layer graphene | high pressure | compression apparatus |
| | monolayer MoS2 | chemical vapor deposition (CVD) | A monolayer MoS sample was grown using chemical vapor deposition (CVD). The lattice coefficient of MoS2 is 0.27 nm. This method can be adapted to use other monolayer two-dimensional | Not specified in the text. | Not specified in the text. | STEM probe, camera length of 20 cm, 80 kV accelerating voltage. |
| | | | materials by changing the two-dimensional lattice parameter used for the analysis. | | | |
| | monolayer graphene | Aberration measurement using Ronchigram | The Ronchigram was recorded at an accelerating voltage of 80 kV from a monolayer graphene sample on a Gatan Oneview camera with 4096 x 4096 pixels using a focused STEM probe. The Ronchigram was divided into local angular areas, and the lateral magnification change in two directions for each segment was measured using an auto-correlation function. The positions of the maxima of three peaks in the auto-correlation | monolayer graphene | accelerating voltage of 80 kV, camera length of 40 cm, focused STEM probe | JEM-ARM300F electron microscope, Gatan Oneview camera, aberration correctors |

| | | | function were located to sub-pixel accuracy with parabolic fits and fitted to an oval function. From these fitted parameters, the aberration coefficients for the probe-forming lens were calculated using the known lattice parameter of graphene and the camera length. | | | |
|---|---|---|---|---|---|---|
| Tin Compensation for the SnS Based Optoelectronic Devices | SnS film | MBE technique | The SnS film was deposited onto a silicon-doped gallium nitride (GaN:Si) layer. The GaN:Si layer was grown on a sapphire substrate and pre-treated by dipping into diluted hydrochloric acid (HCl:H2O = 1:1) for 1 minute to remove the | Hydrochloric acid (HCl), deionized water | Diluted HCl solution (HCl:H2O = 1:1), 1 minute dipping time | MBE chamber, E-beam evaporation technique for Ni electrode deposition, shadow mask with square openings (2 × 2 mm2 ) |

| | | | native oxide layer, followed by rinsing with deionized water before being transferred into the MBE chamber. | | | |
|---|---|---|---|---|---|---|
| Tin Compensation for the SnS Based Optoelectronic Devices | GaN:Si layer | MBE growth | A 2 μm thick GaN:Si window layer was grown on a sapphire substrate. The GaN:Si layer was used as an n-type window electrode in the SnS/GaN:Si heterojunction device. | Silicon-doped gallium nitride (GaN:Si) | Thickness of 2 μm | MBE technique |
| Tin Compensation for the SnS Based Optoelectronic Devices | Ni electrodes | E-beam evaporation | A shadow mask with square openings (2 × 2 mm2) was used to deposit 120 nm thick Ni electrodes by E-beam evaporation technique. | Nickel (Ni) | Thickness of 120 nm | E-beam evaporation technique, shadow mask with square openings (2 × 2 mm2) |
| Tin Compensation for the SnS Based Optoelectronic Devices | SnS | van der Waals epitaxy | Highly textured SnS was grown on mica substrate using a MBE system. | Sn and S (not specified in detail in the main text, but implied by the | Not explicitly mentioned in the main text. | MBE system |

| | | | | context of MBE system usage). | | |
|---|---|---|---|---|---|---|
| Tin Compensation for the SnS Based Optoelectronic Devices | SnS | chemical vapor deposition | SnS nanowires were synthesized via chemical vapor deposition. | Not explicitly mentioned in the main text. | Not explicitly mentioned in the main text. | Chemical vapor deposition system |
| Tin Compensation for the SnS Based Optoelectronic Devices | SnS | congruent thermal evaporation | 3.88% Efficient Tin Sulfide Solar Cells were fabricated using congruent thermal evaporation. | Sn and S (implied by the context of SnS solar cells). | Not explicitly mentioned in the main text. | Thermal evaporation system |
| Tin Compensation for the SnS Based Optoelectronic Devices | SnS | co-evaporation | Tin sulphide thin films were synthesized by co-evaporation. | Sn and S (implied by the context of tin sulphide thin films). | Not explicitly mentioned in the main text. | Co-evaporation system |
| Tin Compensation for the SnS Based Optoelectronic Devices | SnS | heating Se-Sn layers | Semiconducting tin selenide thin films were prepared by heating Se-Sn layers. | Se and Sn | Heating process | Heating apparatus |
| | FeSe monolayer films | Thin film growth on SrTiO3 substrates | Monolayer films of FeSe were grown on 0.05 wt% Nb-doped SrTiO3 substrates. The substrates were degassed for | Fe (99.98%), Se (99.999%) | Degassing at 600 °C for 2 h, annealing at 925 °C for 12 min, growth at 300 °C, post-growth | Knudsen cells, quartz crystal balance, reflection high-energy electron diffraction, ARPES chamber |

| | | | 2 h at 600 °C and then annealed for 12 min at 925 °C. Fe (99.98%) and Se (99.999%) were co-evaporated from Knudsen cells with a flux ratio of 1:10, measured by a quartz crystal balance, and a growth rate of 0.31 unit cell per minute. The growth process was monitored using reflection high-energy electron diffraction. After growth, the FeSe monolayer films were annealed at 350 °C for 20 h and subsequently transferred in situ into the ARPES chamber. | | annealing at 350 °C for 20 h | |
|---|---|---|---|---|---|---|
| | FeSe/STO monolayer | in situ potassium deposition | Starting from a sample annealed at 350°C for 20 | Potassium (K) | Annealing at 350°C for 20 h, in situ | Angle-resolved photoemission |

| | | | | | |
|---|---|---|---|---|---|
| | | h, potassium atoms are deposited in situ onto the surface of the FeSe/STO monolayer to achieve higher electron doping levels. The electron concentration increases from 0.164 electrons per unit cell in the pristine sample to 0.212 and 0.214 after successive rounds of K deposition. The process is monitored using angle-resolved photoemission spectroscopy (ARPES) to observe the evolution of the Fermi surface (FS) and band structure. | | potassium deposition, low temperature (14 K and 70 K) ARPES measurements | spectroscopy (ARPES), He discharge lamp (He Ia and He II lines), scanning tunneling microscopy |
| FeSe bulk crystals | standard synthesis | The superconductivity of FeSe bulk crystals is studied using angle- | Not specified | Low temperature (14 K) ARPES measurements | Angle-resolved photoemission spectroscopy (ARPES) |

| | | | resolved photoemission spectroscopy (ARPES) and compared to the monolayer films. The FS of FeSe bulk crystals is distinct from the monolayer films, lacking the doubly degenerate electron-like pockets at the M point. | | | |
|---|---|---|---|---|---|---|
| | AxFe2 ySe2 | potassium doping | Potassium is used to dope electron carriers into AxFe2 ySe2, leading to a relatively high superconducti ng critical temperature (Tc). The exact mechanism of superconducti vity enhancement is studied using ARPES and band calculations. | Potassium (K) | Electron doping, low temperatur e (14 K) ARPES measurem ents | Angle-resolved photoemissi on spectroscop y (ARPES), band calculations |
| | (Li,Fe)O HFeSe | liquid-gating technique | The superconducti vity of | Not specified | Bulk electron doping via | Angle-resolved photoemissi |

| | (Li,Fe)OHFeSe is enhanced through the use of a liquid-gating technique in the bulk. The material is studied using ARPES to understand the electronic structure and superconducting properties. | liquid-gating, low temperature (14 K) ARPES measurements | on spectroscopy (ARPES) |
|---|---|---|---|